\pdfoutput=1

\documentclass[11pt]{article}

\usepackage[final]{acl}
\usepackage{xcolor}
\usepackage{colortbl}
\usepackage{times}
\usepackage{latexsym}
\usepackage{xspace}
\usepackage[T1]{fontenc}
\usepackage[utf8]{inputenc}
\usepackage{marvosym}
\usepackage{microtype}
\usepackage{booktabs}
\usepackage{inconsolata}
\usepackage{multirow}
\usepackage{amsmath}
\DeclareMathOperator*{\argmin}{arg\,min}
\usepackage{amssymb}
\usepackage{bm}
\usepackage{makecell}
\usepackage[ruled,vlined]{algorithm2e}
\usepackage{graphicx}
\usepackage{tabularx}
\usepackage{enumitem}
\newcolumntype{L}{>{\raggedright\arraybackslash}X} 
\newcolumntype{C}{>{\centering\arraybackslash}X}   
\newcolumntype{R}{>{\raggedleft\arraybackslash}X}  
\newcommand{\headname}{language-specific cross-modal attention heads\xspace}
\newcommand{\shift}{language shift vectors\xspace}
\newcommand{\model}[1]{\textsc{#1}\xspace}
\newcommand{\ours}{\model{CLAIM}}

\definecolor{lightgreen}{HTML}{D9FBD3}
\definecolor{lightgray}{gray}{0.9} 
\setitemize{itemsep=0pt,topsep=0pt,parsep=0pt,partopsep=0pt}
\title{\ours: Mitigating Multilingual Object Hallucination in Large Vision-Language Models with Cross-Lingual Attention Intervention}

\author{Zekai Ye$^{1}$\thanks{~Equal Contribution}, Qiming Li$^{1}$\footnotemark[1], Xiaocheng Feng$^{1,2}$, Libo Qin$^3$, Yichong Huang$^1$, Baohang Li$^1$, \\
\textbf{Kui Jiang$^1$, Yang Xiang$^2$, Zhirui Zhang$^4$, Yunfei Lu$^4$, Duyu Tang$^4$, Dandan Tu$^4$, Bing Qin$^{1,2}$}\\
  $^{1}$Harbin Institute of Technology\quad \quad \quad $^2$Peng Cheng Laboratory\\
  \quad \quad \quad$^{3}$Central South University\quad\quad\quad$^4$Huawei Technologies Co., Ltd\\
  \texttt{\{zkye,qmli,xcfeng\}@ir.hit.edu.cn}
  \\
}

\begin{document}
\maketitle
\begin{abstract}
Large Vision-Language Models (LVLMs) have demonstrated impressive multimodal abilities but remain prone to multilingual object hallucination, with a higher likelihood of generating responses inconsistent with the visual input when utilizing queries in non-English languages compared to English.
Most existing approaches to address these rely on pretraining or fine-tuning, which are resource-intensive. 
In this paper, inspired by observing the disparities in cross-modal attention patterns across languages, we propose \textbf{\underline{C}ross-\underline{L}ingual \underline{A}ttention \underline{I}ntervention for \underline{M}itigating Multilingual Object Hallucination (CLAIM)} in LVLMs, a novel near training-free method by aligning attention patterns. 
\ours first identifies language-specific cross-modal attention heads, then estimates \shift from English to the target language, and finally intervenes in the attention outputs during inference to facilitate cross-lingual visual perception capability alignment. 
Extensive experiments demonstrate that \ours achieves an average improvement of 13.56\% (up to 30\% in Spanish) on the POPE and 21.75\% on the hallucination subsets of the MME benchmark across various languages.
Further analysis reveals that multilingual attention divergence is most prominent in intermediate layers, highlighting their critical role in multilingual scenarios.
\end{abstract}

\section{Introduction}
\label{sec:introduction}

Large Vision-Language Models (LVLMs) 
have made significant strides in bridging visual and textual content \cite{bai2023qwenvl,liu2024visual,ye2023mplug}, leading to notable developments in numerous downstream tasks \cite{shah2023lm,zhu2023prompt,zhang2024vision}. 
However, LVLMs still suffer from serious object hallucination, \textit{i.e.}, generating responses that are inconsistent with the visual input \cite{Yin2023b,Bai2024,Huang2023SurveyHL}, such as misidentifying the presence of objects in an image or providing inaccurate descriptions of their attributes.
This issue becomes even more severe when processing non-English queries \cite{schneider_m5_2024,qu2024mitigating,romero_cvqa_2024}, a challenge referred to as \textbf{multilingual object hallucination} in LVLMs.

Rencent research \cite{qu2024mitigating} focus on mitigating multilingual object hallucination in LVLMs via adopting Supervised Fine-Tuning (SFT) \cite{liu2023mitigating} and Direct Preference Optimization (DPO) \cite{zhao2023hallucinations}.
However, these technologies rely on large-scale annotated image-text datasets, which are extremely expensive, time-consuming, and computation-consuming for non-English especially low-resource languages.
Therefore, it is urgent to develop a training-free method for mitigating multilingual object hallucination, with further understanding of the behavioral disparities of LVLMs in multilingual scenarios.


\begin{figure}
  \centering
  \includegraphics[width=\columnwidth]{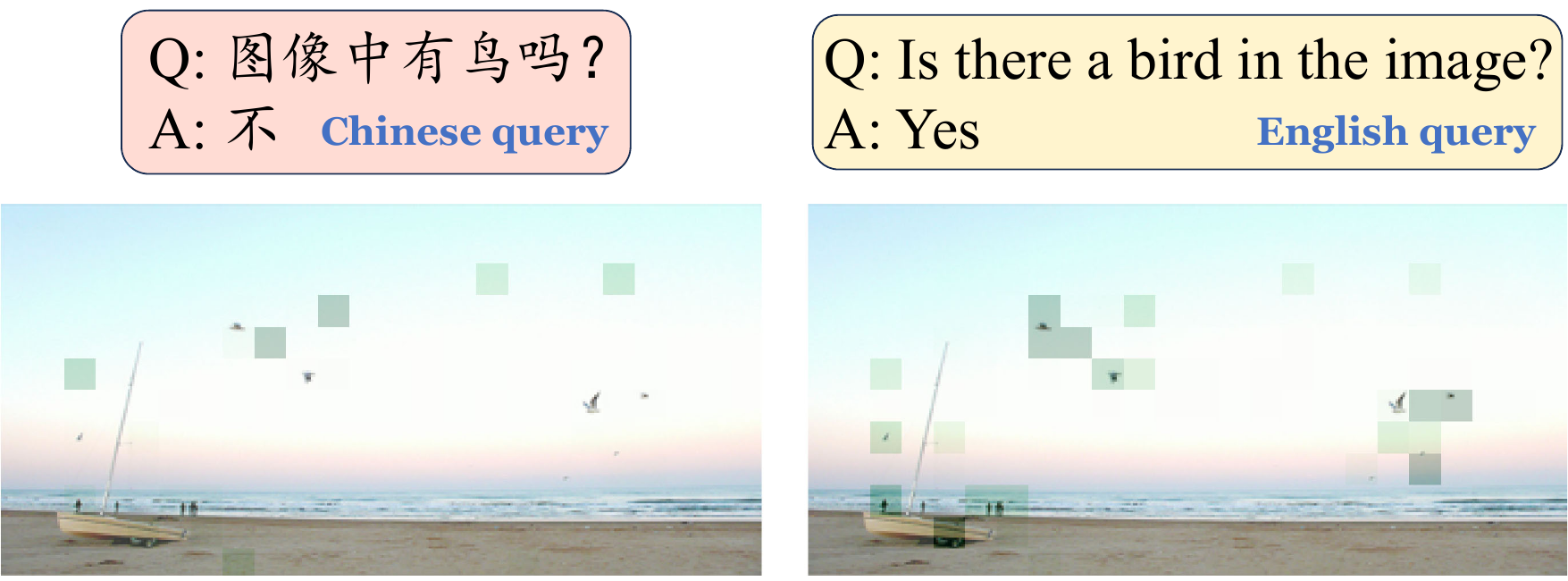}
  \caption{A comparison of attention weights map between Chinese and English query. 
  In English query, LVLM correctly focuses on the key object "\texttt{bird}" in the image, leading to an accurate response. However, in Chinese query, the model exhibits hallucination.}
  \label{figure:intro}
\end{figure}
Inspired by prior research \cite{liu2025paying,bi2024unveiling,chen2024ict,jiang2024devils} highlighting the crucial role of attention in bridging textual and visual information, we discover the significant difference in attention patterns of LVLMs across languages.
Specifically, as illustrated in Figure \ref{figure:intro}, the model pays attention to distinct areas of the image when processing queries in different languages.
Under non-English queries, LVLMs' intermediate layers can even exhibit approximately a 32\% decrease in attention to image regions relevant to queries (\S \ref{sec:ma}).
This finding motivates us to guide the inference process of LVLMs for non-English queries by leveraging the cross-modal attention patterns in English scenarios, as LVLMs are typically well-trained on large English image-text data and perform best in English.

To this end, we propose \textbf{\underline{C}ross-\underline{L}ingual \underline{A}ttention \underline{I}ntervention for \underline{M}itigating Multilingual Object Hallucination (CLAIM)} in LVLMs, a near training-free, plug-and-play method that is applicable during the inference stage.
We first identify \headname, \textit{i.e.}, 
the attention heads behaving quite differently for visual tokens in the same meaning queries across various languages.
Next, we estimate \shift for caption queries of images from English to the target language. During the inference stage, we apply shift vectors to intervene in attention outputs of these heads to align with English visual perception capabilities, reducing the likelihood of multilingual object hallucination. 

Experiments conducted on LLaVA-1.5 \cite{liu2024improved} and Qwen-VL-Chat \cite{bai2023qwenvl} demonstrate that \ours results in an average improvement of $13.56\%$ on the POPE \cite{li2023evaluating} benchmark and $21.75\%$ on the hallucination subsets of the MME \cite{fu2023mme}.

Our contributions are summarized as follows:
\begin{itemize}
\item We reveal significant cross-modal attention divergence across languages in LVLMs.

\item We propose \ours, a novel inference-time method that aligns non-English attention patterns with English, mitigating multilingual object hallucination with low cost. 


\item We analyze LVLMs' attention patterns in multilingual scenarios, highlighting the role of intermediate layers in cross-modal inference.
\end{itemize}
\section{Related Work}

\paragraph{Multilingual Large Vision-Language Models}

Leveraging the advanced capabilities of Large Language Models (LLMs) \cite{Llama2023,chiang2023vicuna,qin2025survey}, Large Vision-Language Models (LVLMs) \cite{yin2023survey,wu2023multimodal} integrate visual encoders \cite{dosovitskiy2020image,radford2021learning} and feature projectors, allowing them to process and generate content from both visual and textual inputs.
Built on the strong multilingual language model LLaMA-2 \cite{touvron2023llama}, which is trained on a diverse multilingual corpus, LLaVA-1.5 \cite{liu2024improved} is inherently a multilingual LVLM.
Qwen-VL-Chat \cite{bai2023qwenvl} is another multilingual LVLM with an English-centric design, trained on a large corpus of Chinese-language data and built upon Qwen \cite{bai2023qwen}.
Existing research \cite{geigle_mblip_2023,andersland2024amharic,maaz_palo_2024} employ training-based approaches to enhance the multilingual capabilities of LVLMs.
Despite significant progress, LVLMs still struggle with multilingual object hallucination, limiting their global applicability in diverse countries and languages.

\paragraph{Hallucination in LVLMs}
LVLMs often generate text outputs that are inconsistent with the visual input, a issue commonly referred to as the hallucination phenomenon.
To mitigate hallucination, some methods focusing on the training phase utilize instruction \cite{liu2023mitigating}, reinforcement learning with human/AI feedback \cite{yu2024rlaif}, or model structure enhancement \cite{chen2024alleviatinghallucinationslargevisionlanguage}. 
Another line of methods \cite{leng2024mitigating,chen2024halc,zhong2024investigating,huang2024opera} reduce the likelihood of hallucination by performing conservative decoding on the original inputs and the inputs with disturbed contents.
However, the above approaches proposed for mitigating hallucination only focus on their effectiveness in English. 
MHR \cite{qu2024mitigating} first attempts to mitigate multilingual object hallucination by SFT and DPO.
In this paper, we propose a novel method for mitigating multilingual object hallucination without datasets construction and training.

\begin{figure*}
  \centering
  \includegraphics[width=\textwidth]{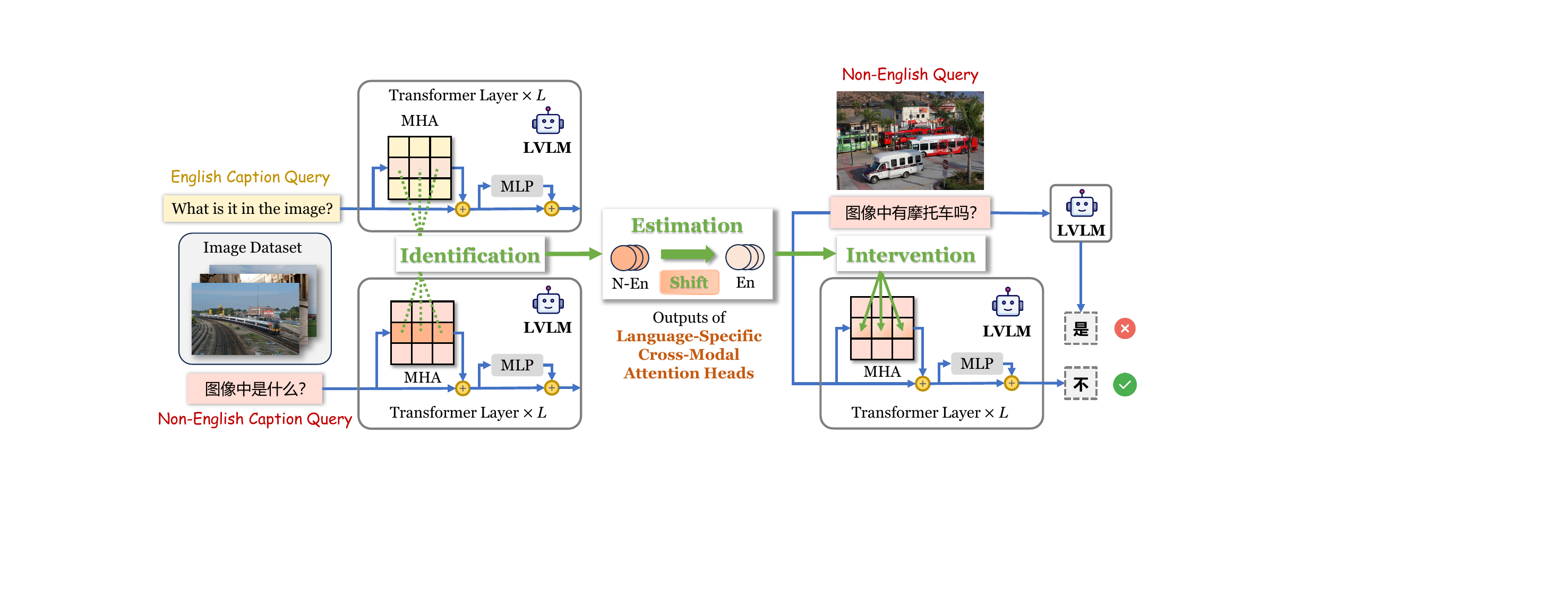}
  \caption{Overview of our proposed \ours method. 
  A block of MHA in the figure represents a attention head.
  \ours intervene in identified \headname using estimated \shift. 
  \textbf{(1) Identification of Language-Specific Cross-Modal Attention Heads \S \ref{sec:Identification}}: We train probes to identify the \headname, which exhibit significantly different behavior across languages associated with visual perception.
\textbf{(2) Estimation of Language Shift Vectors \S \ref{sec:Estimation}}: We estimate the \shift in attention outputs from English to the target non-English language for identical images queried with captions.
\textbf{(3) Intervention during Inference \S \ref{sec:Intervention}}: During inference, we 
apply \shift to intervene in the \headname for mitigating multilingual object hallucination.
  }
  \label{figure:pipeline}
\end{figure*}
\section{Methodology}
In this section, we first introduce the overall process of \ours in Figure \ref{figure:pipeline}, followed by the preliminary for the attention mechanism of LVLMs. 
Then, we describe the three parts of \ours in detail.
\subsection{Preliminary}

Modern LVLMs \cite{bai2023qwenvl,liu2024improved} typically comprise three key components: a visual encoder, a feature projector, and a language decoder.  
Specifically, the visual encoder first transforms the input visual image into visual features. The feature projector then maps these features to the input space of the language decoder, producing the visual embeddings $\bm{P} = \{p_i\}_{i=0}^{n}$, where \(p_i\) represents the visual embedding corresponding to the \(i\)-th image patch, and \(n\) denotes the total number of patches.  
Similarly, the textual input is mapped to textual embeddings \(\bm{T} = \{t_i\}_{i=0}^{m}\), where \(t_i\) corresponds to the \(i\)-th textual token, and \(m\) represents the number of text tokens. The visual and textual embeddings are then concatenated to form the input embeddings \(\bm{X} = [\bm{P}, \bm{T}]\) for the language decoder.  

During the forward pass of the language decoder, the input embeddings \(\bm{X}\) serve as the hidden states for the first self-attention layer \cite{vaswani2017attention}.  
The \(h\)-th attention head in the \(l\)-th self-attention layer applies linear transformations to project the hidden states into queries \(\bm{Q}_h^{l} \in \mathbb{R}^{e \times d}\), keys \(\bm{K}_h^{l} \in \mathbb{R}^{e \times d}\), and values \(\bm{V}_h^{l} \in \mathbb{R}^{e \times d}\). 
Here, \(e=n+m\) and \(d\) denotes the head-specific hidden dimension. The attention scores \(\bm{A}_h^{l} \in \mathbb{R}^{e \times e}\) are then calculated based on \(\bm{Q}_h^{l}\) and \(\bm{K}_h^{l}\) as follows:
\begin{align}
{\widetilde{\bm{A}}^{l}_h} = \textrm{softmax}(&{\bm{A}^{l}_h}+\bm M),
{\bm{A}^{l}_h} = \frac{\bm Q^{l}_h {\bm{K}^{l}_h}^T}{\sqrt{d}},\\
\bm M[i,j]&=\begin{cases} 
0 & \mathrm{if~}j\leq i \\ -\infty & \mathrm{if~}j>i
\end{cases}
\label{eq:attention_weights}
\end{align}
where $\bm M$ is the causal mask matrix. The attention weights $\widetilde{\bm{A}}^{l}_h$ estimate the relevance of each token, which are used to reweight the values $\bm V^{l}_h$  from each token, producing the attention outputs $\bm O^{l}_h \in \mathbb R^{e \times d}$,
\begin{equation}
\bm O^{l}_h =  \widetilde{\bm A}^{l}_h \bm V^{l}_h. \\
\label{eq:attention_output}
\end{equation}
At each layer, the hidden states pass through multi-head attention (MHA), which comprises $H$ independent attention heads, each performing separate linear transformations. 
Specifically, the MHA mechanism can be formulated as:
\begin{equation}
    \bm{X}^{l+1} = \bm{X}^{l} + \sum_{h=1}^{H} \bm O^{l}_h \bm{W}_h^{l},
\end{equation}
where $\bm{W}_h^{l} \in \mathbb{R}^{ d \times Hd}$ maps d-dimensional attention outputs of heads into hidden state representations, which are then fed into a standard multilayer perceptron (MLP) for further processing. 
Finally, the hidden state of the last token is decoded into a next-token prediction distribution.


\subsection{Identification of Language-Specific Cross-Modal Attention Heads}
\label{sec:Identification}

Since LVLMs generate tokens in an auto-regressive manner, our method focuses on the attention 
matrices of the last input token, ${\bm{A}^{l}_h}[e]$, 
which aggregates the most comprehensive visual and textual information.
We mask ${\bm{A}^{l}_h}[e]$ to exclude attention toward all textual tokens, allowing us to identify the \headname, which capture the variations in cross-modal attention patterns where semantically equivalent text in different languages attends to the same image. 
If we do not exclude the attention from the last input token to preceding text tokens, it would introduce substantial text-language-specific features, leading to the identification of attention heads specialized for the input text's language, which diverges from the primary motivation and purpose of our work.
\begin{align}
    \widehat{\bm M}[i,j]=&
    \begin{cases} 
0 & \mathrm{if~}j\leq i \\ -\infty & \mathrm{if~}j>i \mathrm{~or~} (i=e \mathrm{~and~} j>n)
\end{cases} \\
    \widehat{\bm O}^{l}_h &=  \textrm{softmax}(\bm{A}^{l}_h+\widehat{\bm M}) \bm V^{l}_h.
\end{align}

For each image $\bm P_i$, we construct caption queries in both English and the target non-English language. 
A caption query refers to a request where the goal is to generate a textual description (a caption) for a given image which designed to stimulate LVLMs' visual perception capabilities and contributes to mitigating multilingual object hallucination. 
The English query, $T_{en}$, is ``What is it in the image?'', while $T_{tgt}$ is its translation into the target language. 
Then, both queries are fed into the model along with the image for the standard inference process, deriving $\bm x_i \in \left\{ \widehat{\bm O}^{en,l}_{i,h}[e_i], \widehat{\bm O}^{tgt,l}_{i,h}[e_i] \right\}$.

Probe \cite{li2024inference} ${f_h^l}^*$ is a binary
classifier, trained to predict language labels based on $\bm x_i$.
The language labels $y_i \in \left\{1,-1\right\}$ corresponds to English and the target language respectively.
We train probes using $B$ samples for each attention head $H_h^l$.
Finally, we evaluate probes using test samples, identifying \headname for Top-K classification accuracy.
The formulas are summarized as:
\begin{align}
{f_h^l}^* = \argmin_{f_h^l} \sum_{i=1}^{B} \mathcal{L}\left(f_h^l\left(\bm x_i\right), y_i\right),\\
H_s = \{H_{h}^{l} \mid H_{h}^{l} \in TopK(\text{Acc}({f_h^l}^*))\},
\end{align}
where $\mathcal{L}$ is the loss function of probes and $H_s$ is the set of \headname, $K$ denoted as the number of selected heads.

\subsection{Estimation of Language Shift Vectors}
\label{sec:Estimation}
Given the sets $\{(T_{en}, P_i)\}_{i=1}^{B}$ and $\{(T_{tgt}, P_i)\}_{i=1}^{B}$, we derive $\{\bm O^{en,l}_{i,h}\}_{i=1}^{B}$ and $\{\bm O^{tgt,l}_{i,h}\}_{i=1}^{B}$ respectively through the standard inference process of the model, estimating the \shift $\bm{S}_{h}^{l}$:
\begin{equation}
   \bm{S}_{h}^{l} = \frac{1}{B} \sum_{i=1}^{B}\left( \bm O_{i,h}^{en,l}[e_i] - \bm O_{i,h}^{tgt,l}[e_i]\right).
\end{equation}

The shifts estimate the attention disparities between English and the target language for visual perception alignment.
Notably, we do not use $\widehat{\bm O}_{i,h}^{en,l}, \widehat{\bm O}_{i,h}^{tgt,l}$ to estimate \shift as these matrices are not directly derived from the standard inference process. 
Instead, in order to preserve the original representation space of the model, we opt to utilize $\bm O_{i,h}^{en,l}$ and $\bm O_{i,h}^{tgt,l}$ providing more reliable shifts. 
As the visual and textual representations in $\bm O_h^l$ are not orthogonal,
the \shift inherently capture multimodal information understanding which aids in mitigating multilingual object hallucination.

\subsection{Intervention during Inference}
\label{sec:Intervention}
Finally, we apply \shift to intervene in \headname during inference in non-English queries:
\begin{align}
\bm{X}^{l+1} = \bm{X}^{l} &+ \sum_{h=1}^{H} (\bm O^{l}_h+\mathbb{I}_{h}^l\alpha\bm{S}_{h}^{l}) \bm{W}_h^{l}.
\end{align}
$\mathbb{I}_{h}^{l}$ is an indicator function, which is $1$ if $H_h^l \in H_s$ and $0$ otherwise, and $\alpha$ denotes the intensity of the intervention. 
After intervention, LVLMs leverage their strongest English visual perception proficiency even when processing non-English queries.
Since the intervention are pre-computed, \ours hardly incurs additional latency during the inference stage.
We estimate the inference speed and discuss the results in Appendix \ref{sec:latency}.

\section{Experiment}

\subsection{Datasets}
\paragraph{POPE}
POPE \cite{li2023evaluating} is designed to evaluate object hallucination in the VQA paradigm. It queries LVLMs about the presence of specific objects in a given image while maintaining a balanced 1:1 ratio between existent and non-existent objects. 
The benchmark employs three distinct sampling strategies for negative samples - random, popular, adversarial - with their difficulty levels increasing in that order.
POPE integrates data from three major repositories: MSCOCO \cite{lin2014microsoft}, A-OKVQA \cite{schwenk2022okvqa}, and GQA \cite{hudson2019gqa}. 
Evaluation is conducted using accuracy as the primary metric.
As the original POPE benchmark is available only in English, we translating all queries into multiple languages using Google Translate and meticulously refine the translation results to maintain superior benchmark quality.
Since the text is fairly simple, we directly use Google Translate and find it feasible to verify the translation quality, as detailed in Appendix \ref{sec:trans}.
\begin{table*}[!ht]
\centering
\small 
\setlength\tabcolsep{2pt} 
\fontsize{9}{10}\selectfont
\begin{tabularx}{\textwidth}{lllCCCCCCCCCCCCCC}
\toprule
\multirow{2}{*}{\textbf{Dataset}}& \multirow{2}{*}{\textbf{Setup}} & \multirow{2}{*}{\textbf{Method}}& \multicolumn{7}{c}{\textbf{LLaVA-1.5}} & \multicolumn{7}{c}{\textbf{Qwen-VL-Chat}} \\
\cmidrule(lr){4-10} \cmidrule(lr){11-17}
 &  &  & \textbf{En} & \textbf{Zh} & \textbf{Es} & \textbf{Ru} & \textbf{Pt} & \textbf{Bg} &\textbf{Avg.}& \textbf{En}&\textbf{Zh} & \textbf{Es} & \textbf{Ru} & \textbf{Hi} & \textbf{De}&\textbf{Avg.}\\
\midrule
\multirow{9}{*}{\makecell{COCO}} & \multirow{3}{*}{Random} & \textit{Baseline} & {88.50} & 81.00 & 63.03 & 72.33 & 78.97 & 72.23 & 73.51&86.63 &84.57&68.13&76.83&56.97&77.53&72.81 \\
 &  & \textit{VCD} & -& 81.47 & 67.40 & 73.33 & 78.07 & 72.47 &74.55&  -& 84.70&72.17&75.37&48.37&78.27&71.78 \\
&  & \textit{Ours} & -& \cellcolor{lightgreen}86.50 & \cellcolor{lightgreen}87.33 & \cellcolor{lightgreen}87.50 & \cellcolor{lightgreen}85.40 & \cellcolor{lightgreen}80.50 & \cellcolor{lightgreen}85.45& -&\cellcolor{lightgreen}88.03&\cellcolor{lightgreen}86.93&\cellcolor{lightgreen}82.60&\cellcolor{lightgreen}67.57&\cellcolor{lightgreen}88.07&\cellcolor{lightgreen}82.64  \\
  \cmidrule(lr){3-17} 
  & \multirow{3}{*}{Popular} & \textit{Baseline} & {87.43} & 83.07 & 62.93 & 69.20 &	82.03&	71.17	& 73.68&85.67&83.40&68.03&75.73&62.20&77.20&73.31	\\
 &  & \textit{VCD} & -&83.20  & 67.20 &	70.17	& 80.10	&69.47&74.03&-	&82.70&72.70&74.20&57.40&76.70&72.74 \\
&  & \textit{Ours} & -&\cellcolor{lightgreen}88.00 & \cellcolor{lightgreen}87.53 & \cellcolor{lightgreen}84.03 & \cellcolor{lightgreen}86.13 & \cellcolor{lightgreen}80.07 & \cellcolor{lightgreen}85.15&- & \cellcolor{lightgreen}85.47&\cellcolor{lightgreen}85.90&\cellcolor{lightgreen}80.63&\cellcolor{lightgreen}69.23&\cellcolor{lightgreen}86.43&\cellcolor{lightgreen}81.53  \\
  \cmidrule(lr){3-17} 
  & \multirow{3}{*}{Adversarial} & \textit{Baseline} & {85.20}& 73.40 & 62.87 & 66.07	&73.70&	65.93&68.39&83.90&	80.87&67.53&72.80&63.10&75.90&72.04 \\
 &  & \textit{VCD} & -& 74.27	& 67.30	& 66.47	&73.63	&66.33	&69.60&	-&79.40&71.97&70.83&54.73&75.03&70.39 \\
&  & \textit{Ours} & -& \cellcolor{lightgreen}77.67 & \cellcolor{lightgreen}83.27 & \cellcolor{lightgreen}79.43 & \cellcolor{lightgreen}79.67 & \cellcolor{lightgreen}74.57 & \cellcolor{lightgreen}78.92& -&\cellcolor{lightgreen}82.33&\cellcolor{lightgreen}80.70&\cellcolor{lightgreen}77.53&\cellcolor{lightgreen}69.70&\cellcolor{lightgreen}81.50&\cellcolor{lightgreen}78.35  \\
 \midrule
 
 \multirow{9}{*}{\makecell{OKVQA}} & \multirow{3}{*}{Random} & \textit{Baseline} & {91.00} & 76.13 & 63.40 & 70.30 &	75.03&	69.23	&70.82&88.47&85.10&67.70&74.63&59.53&75.17&72.43	 \\
 &  & \textit{VCD} & -&77.93 & 68.17	& 72.17	& 75.67	&70.33	&72.85&-&	84.30&72.23&73.67&49.07&75.33&70.92\\
&  & \textit{Ours} & -&\cellcolor{lightgreen}86.80 & \cellcolor{lightgreen}85.97 & \cellcolor{lightgreen}85.63 & \cellcolor{lightgreen}83.87 & \cellcolor{lightgreen}80.63 & \cellcolor{lightgreen}84.58&- &\cellcolor{lightgreen}88.13 &\cellcolor{lightgreen}86.03&\cellcolor{lightgreen}83.47&\cellcolor{lightgreen}68.83&\cellcolor{lightgreen}86.13&\cellcolor{lightgreen}82.52  \\
  \cmidrule(lr){3-17} 
  & \multirow{3}{*}{Popular} & \textit{Baseline} & {86.97}	& 74.50 &	63.27&	69.40&	77.03&	67.73&	70.39&88.70&85.60&67.90&75.43&58.87&75.27&72.61	 \\
 &  & \textit{VCD} & -&77.57 & 67.10 &68.40 &	77.03	&68.20	&71.66&-	&84.17	&71.57&73.30&52.40&75.63&71.41 \\
&  & \textit{Ours} & -&\cellcolor{lightgreen}85.00 & \cellcolor{lightgreen}83.40 & \cellcolor{lightgreen}81.87 & \cellcolor{lightgreen}84.27 & \cellcolor{lightgreen}77.97 &\cellcolor{lightgreen}82.50& - & \cellcolor{lightgreen}88.10&\cellcolor{lightgreen}85.70&\cellcolor{lightgreen}83.90&\cellcolor{lightgreen}68.50&\cellcolor{lightgreen}86.03&\cellcolor{lightgreen}82.45  \\
  \cmidrule(lr){3-17} 
  & \multirow{3}{*}{Adversarial} & \textit{Baseline} & {79.57}&	64.40&	62.37&	63.93&	67.97&	63.03&	64.34&82.40&79.97&66.80&72.50&60.17&72.30&70.35	 \\
 &  & \textit{VCD} & -&68.97 &	65.17&	63.90 &	68.67&	64.07 &66.16&	 - &	78.77&69.00&69.63&50.80&71.90&68.02	 \\
&  & \textit{Ours} & -&\cellcolor{lightgreen}77.70 & \cellcolor{lightgreen}75.40 & \cellcolor{lightgreen}76.17 & \cellcolor{lightgreen}75.67 & \cellcolor{lightgreen}71.97 &\cellcolor{lightgreen}75.38&  -&\cellcolor{lightgreen}81.10&\cellcolor{lightgreen}77.20&\cellcolor{lightgreen}78.73&\cellcolor{lightgreen}67.73&\cellcolor{lightgreen}78.87&\cellcolor{lightgreen}76.73  \\
 \midrule
 \multirow{9}{*}{\makecell{GQA}} & \multirow{3}{*}{Random} & \textit{Baseline} & 89.47	&77.03	&63.83	& 70.23&	74.87&	68.07&70.81&87.23 &82.53 &73.03 &74.00 &\cellcolor{lightgreen}65.10 &74.30	&73.95\\
 &  & \textit{VCD} & -&78.00	&68.33	& 71.47 &	73.63&	70.20& 72.33&-& 83.63&76.10 &73.43 &55.50 &80.20&73.77\\
&  & \textit{Ours} & -&\cellcolor{lightgreen}84.27 & \cellcolor{lightgreen}85.13 & \cellcolor{lightgreen}83.57 & \cellcolor{lightgreen}82.53  & \cellcolor{lightgreen}82.67 &\cellcolor{lightgreen}83.63&- &\cellcolor{lightgreen}85.17 &\cellcolor{lightgreen}79.93 &\cellcolor{lightgreen}81.90&62.03 &\cellcolor{lightgreen}80.40 & \cellcolor{lightgreen}77.89 \\
  \cmidrule(lr){3-17} 
  & \multirow{3}{*}{Popular} & \textit{Baseline} & 83.90& 	69.60& 64.03& 	64.53& 	71.53& 	63.40&66.62&85.80 &81.90 &72.17 &75.07 &59.20 &70.40&71.93 	 \\
 &  & \textit{VCD} & -&73.60	& 67.50& 	65.23 &71.87& 	66.20 &68.88&- &82.20 &73.73 &74.17 &54.70 &76.87&72.33 	\\
&  & \textit{Ours} & -&\cellcolor{lightgreen}81.57 & \cellcolor{lightgreen}78.50 & \cellcolor{lightgreen}77.93 & \cellcolor{lightgreen}79.60  & \cellcolor{lightgreen}80.57 &\cellcolor{lightgreen}79.63&- & \cellcolor{lightgreen}84.43&\cellcolor{lightgreen}80.60 &\cellcolor{lightgreen}82.60 &\cellcolor{lightgreen}61.30&\cellcolor{lightgreen}78.07&\cellcolor{lightgreen}77.40  \\
  \cmidrule(lr){3-17} 
  & \multirow{3}{*}{Adversarial} & \textit{Baseline} & 81.17& 	63.17& 	63.30	&64.10	& 67.90	& 62.10	& 64.11&82.63&78.97&69.73 &73.07 &61.73 & 71.63 &71.52 \\
 &  & \textit{VCD} & -&67.33	& 66.97	& 63.87	&67.83 & 	63.37 & 65.87&-& 79.57&73.47 & 71.83&54.97 &76.17&71.20 \\
&  & \textit{Ours} & -&\cellcolor{lightgreen}77.57 & \cellcolor{lightgreen}75.93 & \cellcolor{lightgreen}76.27 & \cellcolor{lightgreen}75.23  & \cellcolor{lightgreen}76.47 &\cellcolor{lightgreen}76.29& -&\cellcolor{lightgreen}79.97 &\cellcolor{lightgreen}76.77 &\cellcolor{lightgreen}79.53 &\cellcolor{lightgreen}62.47 &\cellcolor{lightgreen}77.63&\cellcolor{lightgreen}75.27  \\
\midrule
\multirow{15}{*}{MME}&\multirow{3}{*}{Existence}&\textit{Baseline}&190.0&175.0&130.0&145.0&155.0&125.0&146.0&185.0&190.0&135.0&140.0&106.7&195.0&153.3\\
& &\textit{VCD}&-&180.0&130.0&145.0&150.0&128.3&146.7&-&185.0&155.0&155.0&78.30&195.0&153.7\\
& &\textit{Ours}&-&\cellcolor{lightgreen}195.0&\cellcolor{lightgreen}175.0&\cellcolor{lightgreen}185.0&\cellcolor{lightgreen}175.0&\cellcolor{lightgreen}180.0&\cellcolor{lightgreen}182.0&-&\cellcolor{lightgreen}190.0&\cellcolor{lightgreen}155.0&\cellcolor{lightgreen}185.0&\cellcolor{lightgreen}128.3&\cellcolor{lightgreen}195.0&\cellcolor{lightgreen}170.7\\
\cmidrule(lr){3-17} 
&\multirow{3}{*}{Count}&\textit{Baseline}&155.0&70.00&55.00&58.30&80.00&85.00&69.67&150.0&130.0&143.3&113.3&60.00&136.7&116.7\\
& &\textit{VCD}&-&73.30&73.30&53.30&61.70&105.0&73.33&-&140.0&131.7&100.0&80.00&128.3&116.0\\
& &\textit{Ours}&-&\cellcolor{lightgreen}125.0&\cellcolor{lightgreen}130.0&\cellcolor{lightgreen}110.0&\cellcolor{lightgreen}130.0&\cellcolor{lightgreen}135.0&\cellcolor{lightgreen}126.0&-&\cellcolor{lightgreen}148.3&\cellcolor{lightgreen}153.3&\cellcolor{lightgreen}120.0&\cellcolor{lightgreen}111.7&\cellcolor{lightgreen}137.0&\cellcolor{lightgreen}134.0\\
\cmidrule(lr){3-17} 
&\multirow{3}{*}{Color}&\textit{Baseline}&165.0&80.00&135.0&75.00&110.0&80.00&96.00&180.0&170.0&150.0&153.3&\cellcolor{lightgreen}103.3&165.0&148.3\\
& &\textit{VCD}&-&95.00&145.0&85.00&130.0&88.30&108.7&-&150.0&165.0&146.7&93.30&160.0&143.7\\
& &\textit{Ours}&-&\cellcolor{lightgreen}120.0&\cellcolor{lightgreen}155.0&\cellcolor{lightgreen}125.0&\cellcolor{lightgreen}150.0&\cellcolor{lightgreen}108.3&\cellcolor{lightgreen}131.7&-&\cellcolor{lightgreen}170.0&\cellcolor{lightgreen}165.0&\cellcolor{lightgreen}158.3&90.00&\cellcolor{lightgreen}165.0&149.7\cellcolor{lightgreen}\\
\cmidrule(lr){3-17} 
&\multirow{3}{*}{Position}&\textit{Baseline}&118.3&53.30&63.30&55.00&50.00&46.70&53.67&131.7&63.30&\cellcolor{lightgreen}120.0&93.30&45.00&105.0&85.33\\
& &\textit{VCD}&-&48.30&\cellcolor{lightgreen}93.00&60.00&51.70&56.70&61.93&-&\cellcolor{lightgreen}78.30&101.7&101.7&43.30&93.30&83.67\\
& &\textit{Ours}&-&\cellcolor{lightgreen}66.70&78.30&\cellcolor{lightgreen}85.00&\cellcolor{lightgreen}86.70&\cellcolor{lightgreen}58.30&\cellcolor{lightgreen}75.00&-&63.30&116.7&\cellcolor{lightgreen}103.3&\cellcolor{lightgreen}55.00&\cellcolor{lightgreen}106.0&\cellcolor{lightgreen}88.86\\
\cmidrule(lr){3-17} 
&\multirow{3}{*}{Total Scores}&\textit{Baseline}&628.3&378.3&383.3&333.3&395.0&336.7&365.3&646.7&553.3&548.3&500.0&315.0&601.7&503.7\\
& &\textit{VCD}&-&396.7&441.3&343.3&393.3&378.3&390.6&-&553.3&556.7&503.3&295.0&576.7&497.0\\
& &\textit{Ours}&-&\cellcolor{lightgreen}506.7&\cellcolor{lightgreen}538.3&\cellcolor{lightgreen}505.0&\cellcolor{lightgreen}541.7&\cellcolor{lightgreen}481.7&\cellcolor{lightgreen}514.7&-&\cellcolor{lightgreen}571.7&\cellcolor{lightgreen}590.0&\cellcolor{lightgreen}566.6&\cellcolor{lightgreen}385.0&\cellcolor{lightgreen}603.0&\cellcolor{lightgreen}543.3\\
\bottomrule
\end{tabularx}
\caption{Main results on POPE from COCO, OKVQA, GQA and the hallucination subsets of MME. 
}
\label{tab:POPE}
\end{table*}
\begin{figure*}[!h]
\includegraphics[width=\textwidth]{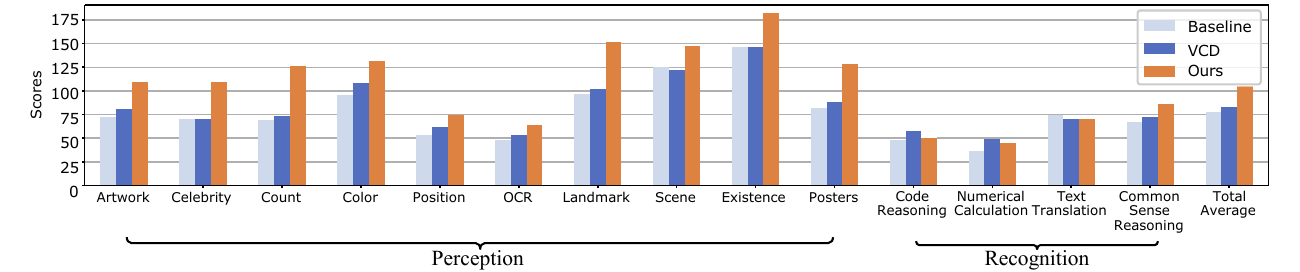}
\caption{Average scores for LLaVA-1.5 across five languages on the MME full dataset.}
\label{fig:mme}
\end{figure*}
\begin{figure*}[t]
\includegraphics[width=\textwidth]{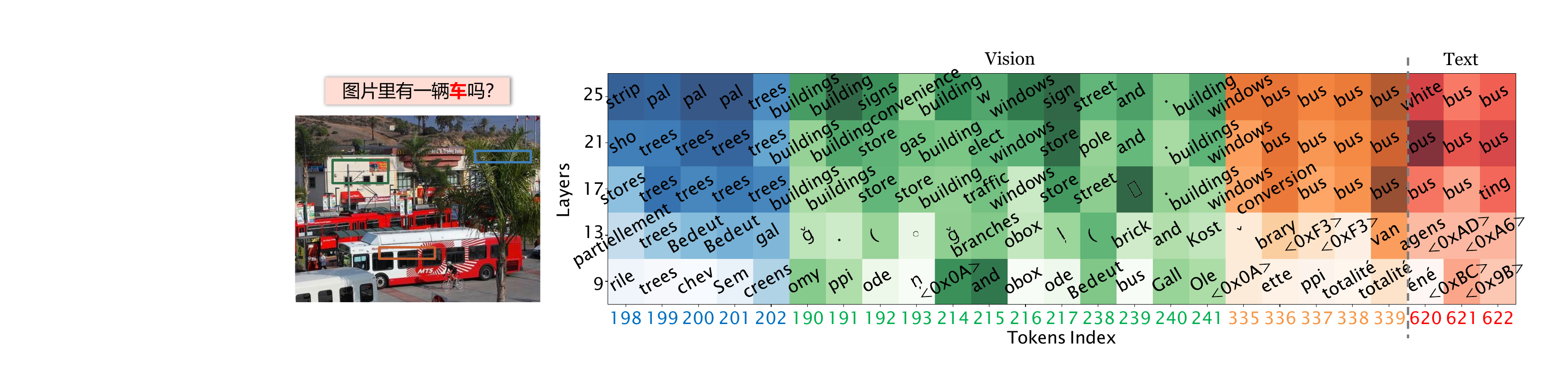}
\caption{Logit lens observation for interpreting LLaVA-1.5 in multilingual scenarios. The depth of block color for the $i$-th token at layer $l$ indicates the magnitude of its contribution to the logits of the final predicted token. The color represents the corresponding tokens in image or text. The query means "Is there a car in the image?".}
\label{fig:logit}
\end{figure*}


\paragraph{MME}
The MME dataset \cite{fu2023mme} serves as a comprehensive benchmark for evaluating LVLMs, 
including 14 subtasks designed to assess both the perceptual and cognitive abilities of LVLMs. 
Performance is measured based on the sum of accuracy scores across individual questions and images.
Following \citet{leng2024mitigating}, in addition to adapting the full dataset, we focus specifically on the existence and count subsets for object-level hallucination evaluation, as well as the position and color subsets for attribute-level assessment. 
Similar to POPE, we translate MME into seven languages using Google Translate and correct the mistakes by human check. 
\subsection{Models and Implementation Details}


Following prior research, we adopt the widely used LLaVA-1.5-7b \cite{liu2024improved} and Qwen-VL-Chat \cite{bai2023qwenvl} as our baseline LVLMs.
More LVLM results can be found in Appendix \ref{sec:llava-next}.
Since no existing training-free methods tailored to mitigating multilingual object hallucination, 
we employ the widely used method, VCD \cite{leng2024mitigating}, as a strong baseline for comparison.
We sample 1,000 images from the COCO-2017 training dataset to complete identification and estimation, discussing the impact of training set size in Appendix \ref{sec:size}.
To determine optimal hyperparameter values for $\alpha$ (intervention intensity) and $K$ (the
number of heads involved in the intervention), we employ a sequential optimization approach. 
Additional details are provided in Appendix \ref{sec:implementation_details}.


\subsection{Main Results}
\paragraph{Results on POPE.}
\textbf{(1) \ours effectively mitigates object-level hallucination} by aligning the strong visual perception capability used for processing English queries with that used for processing non-English queries.
As shown in Table \ref{tab:POPE}, the intervention achieves an average improvement of 17.5\% over the baseline on LLaVA-1.5 and 9.8\% on Qwen-VL-Chat across various languages and settings.
\textbf{(2) \ours enhances performance across both low- and high-resource non-English languages}. This improvement can be attributed to its robust ability to facilitate cross-lingual visual perception capability alignment almost regardless of the language's resource availability.
\textbf{(3) The intervention yields improvements across datasets with different distributions}, suggesting that the intervention represents a generalizable direction to mitigating multilingual object hallucination rather than merely tailored to a specific dataset.
\paragraph{Results on MME.}
This subset extends beyond POPE’s scope, encompassing both object-level and attribute-level hallucinations
\textbf{(1) \ours effectively reduces both object-level and attribute-level hallucination.}
As shown in Table \ref{tab:POPE}, \ours achieves an average improvement of 40.9\% on LLaVA-1.5 and 7.9\% on Qwen-VL-Chat over the baseline across various languages, outperforming VCD.
Specifically, \ours not only mitigates object-level hallucination, as evidenced by the results on the existence and count subsets, but also mitigates attribute-level hallucination, as demonstrated by the color and position subsets.
Detailed results can be found in Appendix \ref{sec:mme_full}.
\textbf{(2) \ours could generally facilitate cross-lingual visual perception capability alignment} by intervening the attention patterns, enabing LVLMs to transfer their English proficiency across various tasks in multilingual queries.
Illustrated in Figure \ref{fig:mme}, the intervention significantly enhances perception-based tasks and generalizes well to cognitive reasoning tasks, as strong image perception serves as the foundation for cognitive processing.
Meanwhile, our method primarily activates attention heads associated with perception, which could unintentionally affect the reasoning pathways of LVLMs.
\begin{table*}[!t]
\centering
\setlength\tabcolsep{2.8pt} 
\fontsize{9}{10}\selectfont
\begin{tabular}{llcccccccccccc}
\toprule
 \multirow{2}{*}{\textbf{Setup}} & \multirow{2}{*}{\textbf{Method}}& \multicolumn{6}{c}{\textbf{LLaVA-1.5}} & \multicolumn{6}{c}{\textbf{Qwen-VL-Chat}} \\
\cmidrule(lr){3-8} \cmidrule(lr){9-14}
&   & \textbf{Zh} & \textbf{Es} & \textbf{Ru} & \textbf{Pt} & \textbf{Bg} &\textbf{Avg.}&\textbf{Zh} & \textbf{Es} & \textbf{Ru} & \textbf{Hi} & \textbf{De}&\textbf{Avg.}\\
\midrule
\multirow{4}{*}{Random} & \textit{Baseline} & 81.00 & 63.03 & 72.33 & 78.97 & 72.23 & 73.51 &84.57&68.13&76.83&56.97&77.53&72.81 \\
& \textit{Mono-Shift} &\cellcolor{lightgreen}86.50&86.70&87.40&\cellcolor{lightgray}84.50&76.77&84.37&\cellcolor{lightgreen}88.03&74.87&80.67&\cellcolor{lightgray}67.27&83.03&78.77  \\
& \textit{Multi-Shift} & \cellcolor{lightgray}85.80&\cellcolor{lightgray}87.00&\cellcolor{lightgreen}87.87&83.60&\cellcolor{lightgray}79.80&\cellcolor{lightgray}84.81&\cellcolor{lightgray}86.70&\cellcolor{lightgray}86.40&\cellcolor{lightgray}81.37&63.30&\cellcolor{lightgray}87.97&\cellcolor{lightgray}81.15  \\
& \textit{Specific-Shift} & \cellcolor{lightgreen}86.50 & \cellcolor{lightgreen}87.33 & \cellcolor{lightgray}87.50 & \cellcolor{lightgreen}85.40 & \cellcolor{lightgreen}80.50 & \cellcolor{lightgreen}85.45&\cellcolor{lightgreen}88.03&\cellcolor{lightgreen}86.93&\cellcolor{lightgreen}82.60&\cellcolor{lightgreen}67.57&\cellcolor{lightgreen}88.07&\cellcolor{lightgreen}82.64  \\
  \cmidrule(lr){2-14} 
\multirow{4}{*}{Popular} & \textit{Baseline} & 83.07 & 62.93 & 69.20 &	82.03&	71.17	& 73.68&83.40&68.03&75.73&62.20&77.20	&73.31\\
& \textit{Mono-Shift} &\cellcolor{lightgreen}88.00&86.90&83.00&85.30&79.10&84.46&\cellcolor{lightgreen}85.47&74.77&78.80&68.07&81.73&77.77  \\
& \textit{Multi-Shift} &\cellcolor{lightgray}87.67&\cellcolor{lightgray}87.23&\cellcolor{lightgray}83.33&\cellcolor{lightgreen}86.17&\cellcolor{lightgray}79.13&\cellcolor{lightgray}84.71&\cellcolor{lightgray}84.30&\cellcolor{lightgray}85.03&\cellcolor{lightgray}80.37&\cellcolor{lightgray}68.93&\cellcolor{lightgray}85.43 &\cellcolor{lightgray}80.81 \\
& \textit{Specific-Shift} &\cellcolor{lightgreen}88.00 &\cellcolor{lightgreen} 87.53 & \cellcolor{lightgreen}84.03 & \cellcolor{lightgray}86.13 & \cellcolor{lightgreen}80.07 & \cellcolor{lightgreen}85.15 & \cellcolor{lightgreen}85.47&\cellcolor{lightgreen}85.90&\cellcolor{lightgreen}80.63&\cellcolor{lightgreen}69.23&\cellcolor{lightgreen}86.43&\cellcolor{lightgreen}81.53  \\
  \cmidrule(lr){2-14} 
\multirow{4}{*}{Adversarial} & \textit{Baseline}& 73.40 & 62.87 & 66.07	&73.70&	65.93	&68.39&	80.87&67.53&72.80&63.10&75.90&72.01 \\
& \textit{Mono-Shift} &\cellcolor{lightgreen}77.67&\cellcolor{lightgray}82.63&78.13&\cellcolor{lightgray}78.33&72.60&77.87&\cellcolor{lightgreen}82.33&73.83&76.00&69.30&79.57&76.21 \\
& \textit{Multi-Shift} &\cellcolor{lightgray}76.70&82.13&\cellcolor{lightgreen}79.53&78.10&\cellcolor{lightgreen}74.83&\cellcolor{lightgray}78.66&\cellcolor{lightgray}81.70&\cellcolor{lightgray}80.50&\cellcolor{lightgray}76.57&\cellcolor{lightgreen}69.97&\cellcolor{lightgray}81.17&\cellcolor{lightgray}77.98  \\
& \textit{Specific-Shift} & \cellcolor{lightgreen}77.67 & \cellcolor{lightgreen}83.27 & \cellcolor{lightgray}79.43 & \cellcolor{lightgreen}79.67 & \cellcolor{lightgray}74.57 & \cellcolor{lightgreen}78.92 &\cellcolor{lightgreen}82.33&\cellcolor{lightgreen}80.70&\cellcolor{lightgreen}77.53&\cellcolor{lightgray}69.70&\cellcolor{lightgreen}81.50&\cellcolor{lightgreen}78.35  \\
\bottomrule
\end{tabular}
\caption{Evaluation results of different intervention estimation approaches on POPE-COCO. \colorbox{lightgreen}{Green} means the best perfomance while \colorbox{lightgray}{gray} means the second-best results.}

\label{tab:Mul}
\end{table*}
\section{Analysis and Discussion}
\subsection{Multilingual Attention Differences}
\label{sec:ma}
As shown in Figure \ref{figure:intro}, when queried in different languages about an object's presence, the model exhibits distinct attention patterns across image regions.
In order to quantitatively analyze multilingual cross-modal attention differences, we conduct a statistical experiment to validate our motivation. 
Specifically, using bounding box annotations from the POPE-COCO dataset, we localize object regions and compute the sum of attention weights within ground-truth bounding boxes for queries in different languages, obtaining $A_b(l,\mathcal{B})$.
\begin{equation}
A_b(l,\mathcal{B}) =  \frac{n}{H|\mathcal{B}|} \cdot \sum_{h=1}^{H} \left( \frac{\sum_{j\in\mathcal{B}} \widetilde{A}_{h}^{l}(e,j)}{\sum_{j=1}^{n} \widetilde{A}_{h}^{l}(e,j)} \right)
\end{equation}
$\mathcal{B}$ denotes the set of patches included in bounding boxes.
The first fraction serves as a normalization factor, eliminating the influence of varying bounding box sizes associated with different queries and averaging $H$ attention heads in each layer.
\begin{figure}[h]
  \includegraphics[width=\columnwidth]{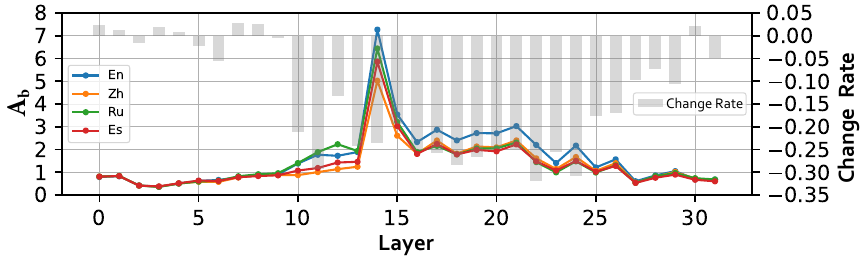}
  \caption{$A_b$ per layer of LLaVA-1.5 across four languages, and the per-layer average change rate of non-English languages $A_b$ relative to English.}
  \label{ab}
\end{figure}

As shown in Figure \ref{ab}, in the initial layers of the model (before layer 10), the model pays little attention to image regions related to queries, and there is almost no difference between multilingual queries, indicating that the model is performing basic global understanding of the image at this stage. 
In the middle layers (10-25), the attention to key regions in the image for non-English languages decreases significantly compared to English, suggesting that the model exhibits disparities in perceptual capabilities under different language queries. 
Notably, a peak appears at layer 14, even though the model's total attention weights for visual tokens are low at this stage (shown in Figure \ref{fig:attn} (a)). 
This may indicate that the model has noticed image regions relevant to queries during its internal reasoning process, performing fine-grained perception.
\subsection{Multilingual Inference in LVLMs}
In order to better illustrate the mechanism behind \ours and improve the interpretability of LVLMs in multilingual scenarios,
by leveraging the logit lens \cite{nostalgebraist2020logitlens} method, which aims to decode the hidden states of the language decoder at various layers, we investigate the internal inference mechanism of LVLMs and uncover how they process and integrate multimodal information, particularly in non-English queries.
The internal inference process of LLaVA-1.5 is illustrated in Figure \ref{fig:logit} as a case study.

For multilingual VQA tasks, English-centric LVLMs face the dual challenge of not only bridging the modality gap between visual and textual information but also mapping non-English queries into the English semantic space to ensure accurate responses.
\textbf{In middle layers, visual tokens are often decoded into their corresponding English concepts}. Similarly, when processing Chinese queries, the model maps them to the English semantic space at these intermediate layers such as the layer 17, consistent with the findings \cite{wendler2024llamas} in multilingual LLMs.
During pretraining, the alignment of vision-text modality relies heavily on English corpora, which guides LVLMs toward interpreting images through an English-centric pathway. 
\textbf{LVLMs interpret the important entities in the query based on the critical information from the image at intermediate stage}. 
The original semantic meaning of "car" in the query (such as the 620th token) is enriched to "bus" under the influence of the image information at the layer 21.

\subsection{Analysis of Intervention}
In order to elucidate the underlying reasons for the efficacy of \ours and to substantiate the robustness of our methodology, we conduct a further investigation with two questions.

\begin{figure}[h]
  \includegraphics[width=\columnwidth]{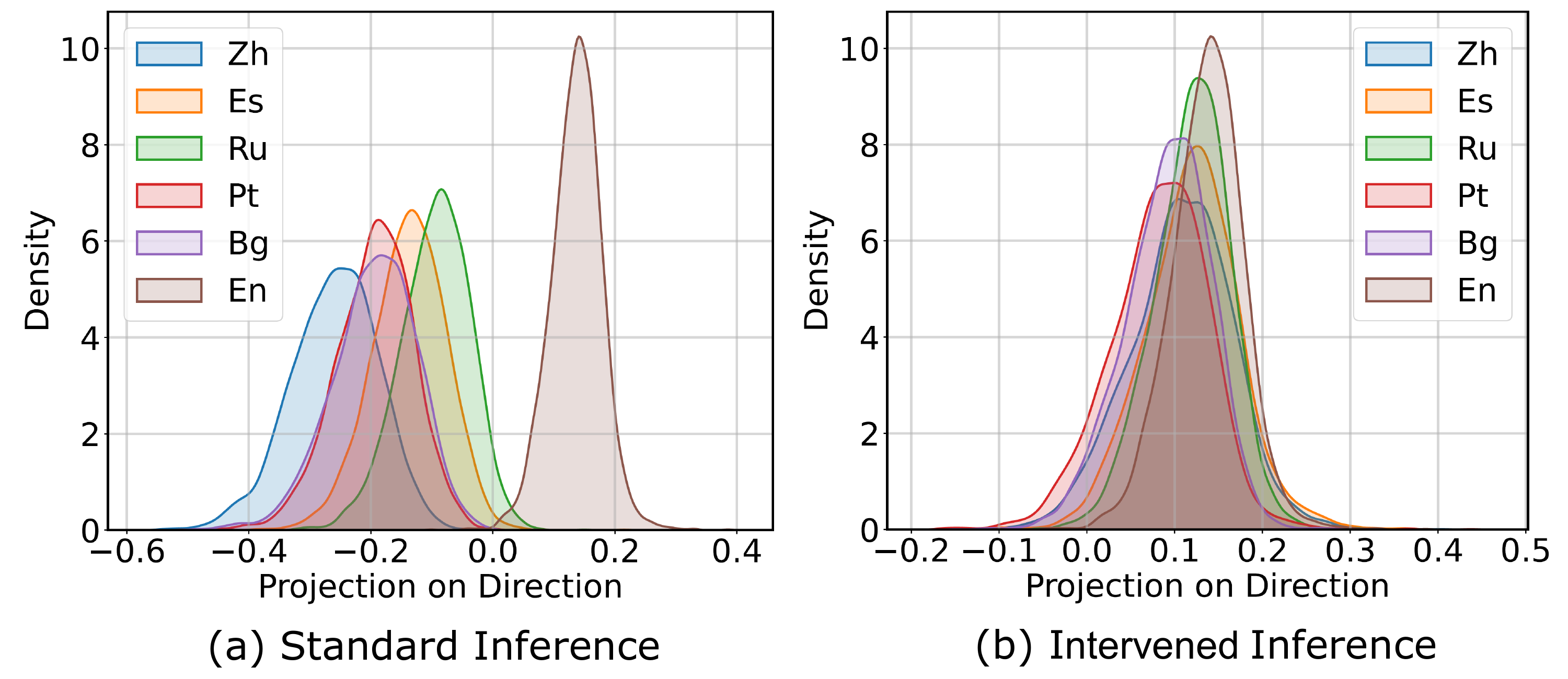}
  \caption{Kernel density estimate plot of cross-modal attention outputs across languages of LLaVA-1.5 before and after multilingual intervention.
  The x-axis represents the inner product between attention outputs and the normal vector of the hyperplane, while the y-axis indicates the density of samples occurring at x value.}
  \label{fig:kda}
\end{figure}

\textit{Does the intervention truly facilitate the cross-lingual alignment of cross-modal attention distribution?}
Illustrated in Figure \ref{fig:kda}, under standard inference, the projection values of English and non-English vectors are distinctly separated at the zero point, indicating that the cross-modal attention patterns of LVLMs exhibit significant misalignment for identical images depending on the language used in the same meaning query.
After intervention, the distributions of non-English languages shift closer to that of English, with more pronounced density peaks. 
This alignment supports the hypothesis that \ours effectively mitigates multilingual object hallucination by reinforcing cross-lingual consistency in visual perception attention.


\textit{Does a unified multilingual intervention (\textit{Multi-Shift}) work?}
In our main experiment, we pair English with each non-English language individually, estimating the \textit{Specific-Shift} for each pair and selecting specialized attention heads to precisely align each non-English languages with English proficiency. 
To examine the impact of multilingual interactions, we construct a mixed set that includes all non-English languages paired with English, estimating the \textit{Multi-Shift} between English and the entire non-English group. 
As shown in Table \ref{tab:Mul}, the \textit{Multi-Shift} demonstrates moderate improvements, even outperforming \ours in some subsets by integrating shared linguistic features.
Additionally, to further validate the generalizability of the intervention, we apply the \textit{Mono-Shift} intervention, derived from the English-Chinese pair, to other languages and observe performance improvements.
It indicates that the cross-lingual intervention can generalize well to unseen non-English languages.
\subsection{Analysis of Attention Heads}
\begin{figure}[h]
  \includegraphics[width=\columnwidth]{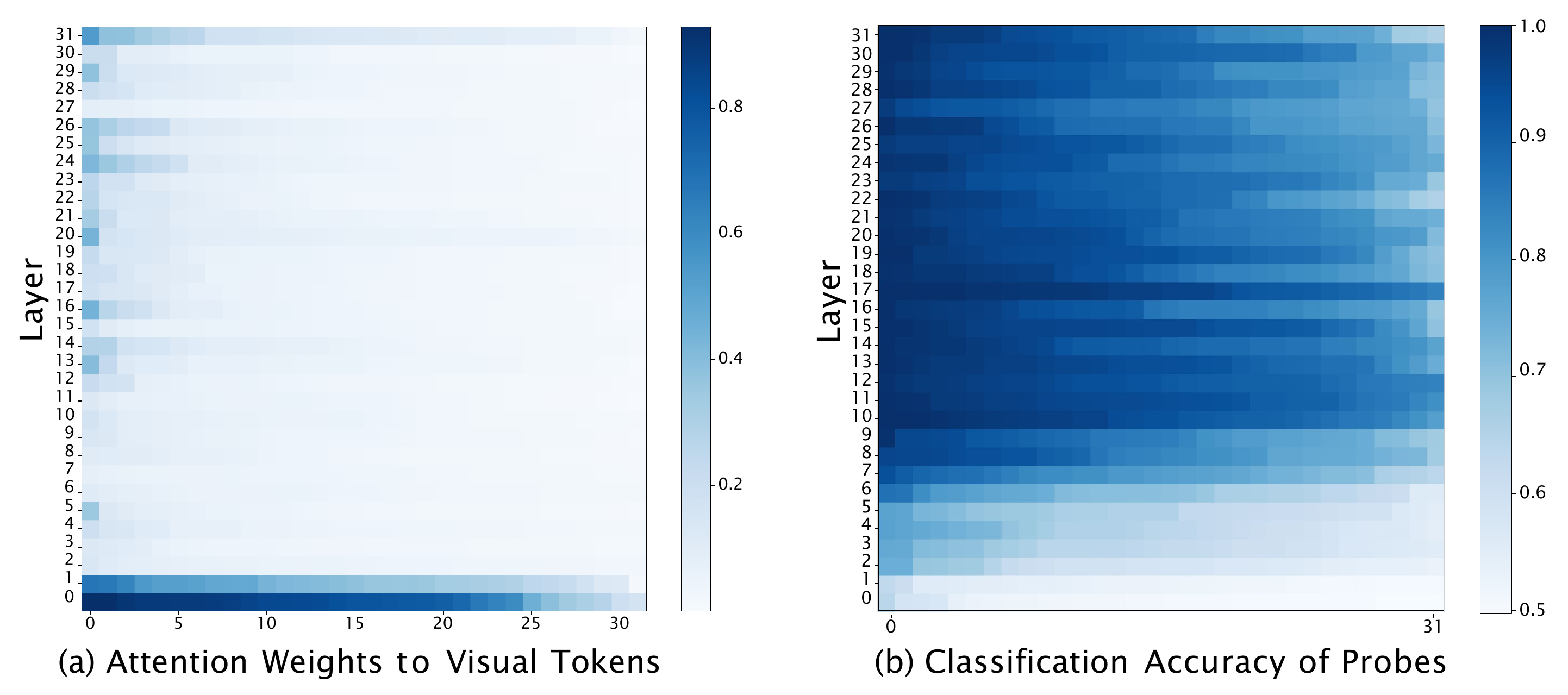}
  \caption{Heatmap of (a) sum attention weights of the last input token toward all visual tokens and (b) the classification accuracy of probes of LLaVA-1.5 across all 32x32 multi-heads, sorted row-wise by value.}
  \label{fig:attn}
\end{figure}

\textbf{LVLMs extract and interpret cross-modal information within \headname}.
As shown in Figure \ref{fig:attn} (b), theses attention heads are predominantly located in the intermediate layers of the model, particularly at layers 10-17, suggesting that cross-modal integration occurs primarily at this stage.
Beyond this, we examine how different layers influence the final prediction. In Figure \ref{fig:logit}, we observe that around layer 21, visual and textual tokens—such as the 339th and 620th tokens, which correspond to the semantics of "bus"—have a significant impact on the logits of the final prediction. Notably, attention heads in these layers exhibit high classification accuracy, indicating their crucial role in linguistic visual perception and understanding.
Conversely, Figure \ref{fig:attn} (a) shows that in the early layers (0 and 1), the attention weights of the last input token to visual tokens are strongest.
However, classification accuracy remains very low in these heads, suggesting that while LVLMs engage in basic image feature extraction at the initial layers, they contribute minimally to linguistic understanding. 
\subsection{Impact of Hyperparameters}
\begin{figure}[h]
  \includegraphics[width=\columnwidth]{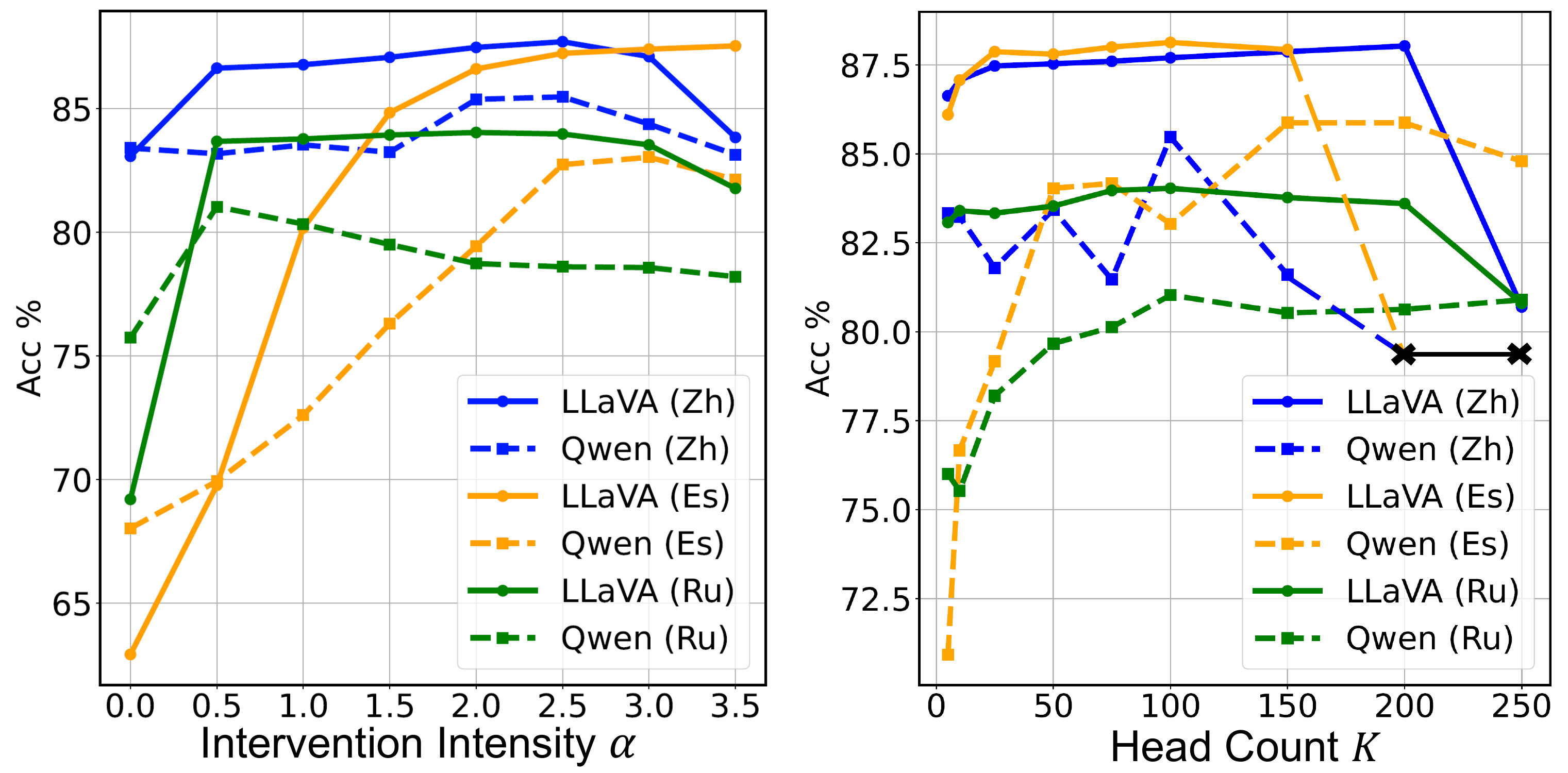}
  \caption{Impact of hyperparameters $\alpha$ and $K$ on the Accuracy for LLaVA-1.5 and Qwen-VL-Chat on the POPE-COCO popular subset. The "x" symbol indicates a value extraordinarily lower than normal.}
  \label{fig:ablation_alpha}
\end{figure}

\ours is primarily governed by two key hyperparameters: the intervention intensity $\alpha$ and the number of heads $K$ involved in the intervention. 
Illustrated in Figure \ref{fig:ablation_alpha}, the ablation experiments vary one parameter while keeping the other fixed, yielding several key insights.
When $\alpha$ is too small, the intervention is insufficient, resulting in suboptimal improvements. 
However, an excessively large $\alpha$ imposes an overly strong intervention, disrupting the LVLMs' capabilities.
For the hyperparameter $K$, we observe that a small $K$ leads to inadequate intervention in \headname, reducing effectiveness.
On the other hand, a large $K$ introduces unnecessary interference by affecting attention heads that encode irrelevant information, ultimately degrading performance.
Overall, \ours achieves performance improvements across a wide range of hyperparameter settings, demonstrating strong robustness to hyperparameter selection.
\section{Conclusion}
In this paper, we propose \textbf{\underline{C}ross-\underline{L}ingual \underline{A}ttention \underline{I}ntervention for \underline{M}itigating Multilingual Object Hallucination (CLAIM)} in LVLMs, a near training-free method that aligns attention patterns across languages. 
Extensive evaluations on POPE and MME benchmarks demonstrate that \ours effectively mitigates multilingual object hallucination and generalizes well across languages and datasets. 
Further analysis reveals that attention discrepancies primarily occur in intermediate layers and LVLMs extract and interpret cross-modal information within \headname., providing deeper insights into multilingual LVLMs inference pathways.

\section{Limitations}
\ours requires distinguishing text and vision information within the attention mechanism to identify \headname, making it applicable only to LVLMs that treat visual and textual tokens equally in the language decoder.
Additionally, our method requires access to the internal layers and representations of LVLMs, limiting its applicability to closed-source models.
How to mitigate multilingual object hallucination as a plug-and-play tool for all LLMs, including those with restricted access, requires further investigation. 
Besides, since \ours aligns non-English attention patterns with English patterns, it may inadvertently reinforce English-centric biases rather than fostering truly multicultural comprehension.

\section{Acknowledgments}
Xiaocheng Feng is the corresponding author of this work. We thank the anonymous reviewers for their insightful comments. This work was supported by the National Natural Science Foundation of China (NSFC) (grant 62276078, U22B2059), the Key R\&D Program of Heilongjiang via grant 2022ZX01A32,  and the Fundamental Research Funds for the Central Universities (Grant No.HIT.OCEF.2023018).
\bibliography{custom}

\begin{thebibliography}{50}
\providecommand{\natexlab}[1]{#1}

\bibitem[{Andersland(2024)}]{andersland2024amharic}
Michael Andersland. 2024.
\newblock Amharic llama and llava: Multimodal llms for low resource languages.
\newblock \emph{arXiv preprint arXiv:2403.06354}.

\bibitem[{Bai et~al.(2023{\natexlab{a}})Bai, Bai, Chu, Cui, Dang, Deng, Fan, Ge, Han, Huang et~al.}]{bai2023qwen}
Jinze Bai, Shuai Bai, Yunfei Chu, Zeyu Cui, Kai Dang, Xiaodong Deng, Yang Fan, Wenbin Ge, Yu~Han, Fei Huang, et~al. 2023{\natexlab{a}}.
\newblock Qwen technical report.
\newblock \emph{arXiv preprint arXiv:2309.16609}.

\bibitem[{Bai et~al.(2023{\natexlab{b}})Bai, Bai, Yang, Wang, Tan, Wang, Lin, Zhou, and Zhou}]{bai2023qwenvl}
Jinze Bai, Shuai Bai, Shusheng Yang, Shijie Wang, Sinan Tan, Peng Wang, Junyang Lin, Chang Zhou, and Jingren Zhou. 2023{\natexlab{b}}.
\newblock Qwen-vl: A frontier large vision-language model with versatile abilities.
\newblock \emph{arXiv preprint arXiv:2308.12966}.

\bibitem[{Bi et~al.(2024)Bi, Guo, Tang, Wen, Liu, and Xu}]{bi2024unveiling}
Jing Bi, Junjia Guo, Yunlong Tang, Lianggong~Bruce Wen, Zhang Liu, and Chenliang Xu. 2024.
\newblock Unveiling visual perception in language models: An attention head analysis approach.
\newblock \emph{arXiv preprint arXiv:2412.18108}.

\bibitem[{Chen et~al.(2024{\natexlab{a}})Chen, Lyu, Gao, Song, and Shen}]{chen2024alleviatinghallucinationslargevisionlanguage}
Beitao Chen, Xinyu Lyu, Lianli Gao, Jingkuan Song, and Heng~Tao Shen. 2024{\natexlab{a}}.
\newblock \href {https://arxiv.org/abs/2405.15356} {Alleviating hallucinations in large vision-language models through hallucination-induced optimization}.
\newblock \emph{Preprint}, arXiv:2405.15356.

\bibitem[{Chen et~al.(2024{\natexlab{b}})Chen, Zhang, Huang, Niu, Zhang, Wen, and Hu}]{chen2024ict}
Junzhe Chen, Tianshu Zhang, Shiyu Huang, Yuwei Niu, Linfeng Zhang, Lijie Wen, and Xuming Hu. 2024{\natexlab{b}}.
\newblock Ict: Image-object cross-level trusted intervention for mitigating object hallucination in large vision-language models.
\newblock \emph{arXiv preprint arXiv:2411.15268}.

\bibitem[{Chen et~al.(2024{\natexlab{c}})Chen, Zhao, Luo, Yao, Li, and Zhou}]{chen2024halc}
Zhaorun Chen, Zhuokai Zhao, Hongyin Luo, Huaxiu Yao, Bo~Li, and Jiawei Zhou. 2024{\natexlab{c}}.
\newblock Halc: Object hallucination reduction via adaptive focal-contrast decoding.
\newblock \emph{arXiv preprint arXiv:2403.00425}.

\bibitem[{Chiang et~al.(2023)Chiang, Li, Lin, Sheng, Wu, Zhang, Zheng, Zhuang, Zhuang, Gonzalez et~al.}]{chiang2023vicuna}
Wei-Lin Chiang, Zhuohan Li, Zi~Lin, Ying Sheng, Zhanghao Wu, Hao Zhang, Lianmin Zheng, Siyuan Zhuang, Yonghao Zhuang, Joseph~E Gonzalez, et~al. 2023.
\newblock Vicuna: An open-source chatbot impressing gpt-4 with 90\%* chatgpt quality.
\newblock \emph{See https://vicuna. lmsys. org (accessed 14 April 2023)}, 2(3):6.

\bibitem[{Cortes(1995)}]{cortes1995support}
Corinna Cortes. 1995.
\newblock Support-vector networks.
\newblock \emph{Machine Learning}.

\bibitem[{Dosovitskiy(2020)}]{dosovitskiy2020image}
Alexey Dosovitskiy. 2020.
\newblock An image is worth 16x16 words: Transformers for image recognition at scale.
\newblock \emph{arXiv preprint arXiv:2010.11929}.

\bibitem[{Fu et~al.(2023)Fu, Chen, Shen, Qin, Zhang, Lin, Yang, Zheng, Li, Sun et~al.}]{fu2023mme}
Chaoyou Fu, Peixian Chen, Yunhang Shen, Yulei Qin, Mengdan Zhang, Xu~Lin, Jinrui Yang, Xiawu Zheng, Ke~Li, Xing Sun, et~al. 2023.
\newblock Mme: A comprehensive evaluation benchmark for multimodal large language models.
\newblock \emph{arXiv preprint arXiv:2306.13394}.

\bibitem[{Geigle et~al.(2023)Geigle, Jain, Timofte, and Glavas}]{geigle_mblip_2023}
Gregor Geigle, Abhay Jain, Radu Timofte, and Goran Glavas. 2023.
\newblock \href {https://doi.org/10.48550/ARXIV.2307.06930} {{mBLIP}: {Efficient} {Bootstrapping} of {Multilingual} {Vision}-{LLMs}}.
\newblock \emph{CoRR}, abs/2307.06930.
\newblock ArXiv: 2307.06930.

\bibitem[{Huang et~al.(2023)Huang, Yu, Ma, Zhong, Feng, Wang, Chen, Peng, Feng, Qin, and Liu}]{Huang2023SurveyHL}
Lei Huang, Weijiang Yu, Weitao Ma, Weihong Zhong, Zhangyin Feng, Haotian Wang, Qianglong Chen, Weihua Peng, Xiaocheng Feng, Bing Qin, and Ting Liu. 2023.
\newblock \href {https://arxiv.org/abs/2311.05232} {A survey on hallucination in large language models: Principles, taxonomy, challenges, and open questions}.
\newblock \emph{Preprint}, arXiv:2311.05232.

\bibitem[{Huang et~al.(2024)Huang, Dong, Zhang, Wang, He, Wang, Lin, Zhang, and Yu}]{huang2024opera}
Qidong Huang, Xiaoyi Dong, Pan Zhang, Bin Wang, Conghui He, Jiaqi Wang, Dahua Lin, Weiming Zhang, and Nenghai Yu. 2024.
\newblock Opera: Alleviating hallucination in multi-modal large language models via over-trust penalty and retrospection-allocation.
\newblock In \emph{Proceedings of the IEEE/CVF Conference on Computer Vision and Pattern Recognition}, pages 13418--13427.

\bibitem[{Hudson and Manning(2019)}]{hudson2019gqa}
Drew~A Hudson and Christopher~D Manning. 2019.
\newblock Gqa: A new dataset for real-world visual reasoning and compositional question answering.
\newblock In \emph{Proceedings of the IEEE/CVF conference on computer vision and pattern recognition}, pages 6700--6709.

\bibitem[{Jiang et~al.(2024)Jiang, Chen, Zhu, Luo, Shen, and Yang}]{jiang2024devils}
Zhangqi Jiang, Junkai Chen, Beier Zhu, Tingjin Luo, Yankun Shen, and Xu~Yang. 2024.
\newblock Devils in middle layers of large vision-language models: Interpreting, detecting and mitigating object hallucinations via attention lens.
\newblock \emph{arXiv preprint arXiv:2411.16724}.

\bibitem[{Leng et~al.(2024)Leng, Zhang, Chen, Li, Lu, Miao, and Bing}]{leng2024mitigating}
Sicong Leng, Hang Zhang, Guanzheng Chen, Xin Li, Shijian Lu, Chunyan Miao, and Lidong Bing. 2024.
\newblock Mitigating object hallucinations in large vision-language models through visual contrastive decoding.
\newblock In \emph{Proceedings of the IEEE/CVF Conference on Computer Vision and Pattern Recognition}, pages 13872--13882.

\bibitem[{Li et~al.(2024)Li, Patel, Vi{\'e}gas, Pfister, and Wattenberg}]{li2024inference}
Kenneth Li, Oam Patel, Fernanda Vi{\'e}gas, Hanspeter Pfister, and Martin Wattenberg. 2024.
\newblock Inference-time intervention: Eliciting truthful answers from a language model.
\newblock \emph{Advances in Neural Information Processing Systems}, 36.

\bibitem[{Li et~al.(2023)Li, Du, Zhou, Wang, Zhao, and Wen}]{li2023evaluating}
Yifan Li, Yifan Du, Kun Zhou, Jinpeng Wang, Xin Zhao, and Ji-Rong Wen. 2023.
\newblock Evaluating object hallucination in large vision-language models.
\newblock In \emph{The 2023 Conference on Empirical Methods in Natural Language Processing}.

\bibitem[{Lin et~al.(2014)Lin, Maire, Belongie, Hays, Perona, Ramanan, Doll{\'a}r, and Zitnick}]{lin2014microsoft}
Tsung-Yi Lin, Michael Maire, Serge Belongie, James Hays, Pietro Perona, Deva Ramanan, Piotr Doll{\'a}r, and C~Lawrence Zitnick. 2014.
\newblock Microsoft coco: Common objects in context.
\newblock In \emph{Computer Vision--ECCV 2014: 13th European Conference, Zurich, Switzerland, September 6-12, 2014, Proceedings, Part V 13}, pages 740--755. Springer.

\bibitem[{Liu et~al.(2023)Liu, Lin, Li, Wang, Yacoob, and Wang}]{liu2023mitigating}
Fuxiao Liu, Kevin Lin, Linjie Li, Jianfeng Wang, Yaser Yacoob, and Lijuan Wang. 2023.
\newblock Mitigating hallucination in large multi-modal models via robust instruction tuning.
\newblock In \emph{The Twelfth International Conference on Learning Representations}.

\bibitem[{Liu et~al.(2024{\natexlab{a}})Liu, Li, Li, and Lee}]{liu2024improved}
Haotian Liu, Chunyuan Li, Yuheng Li, and Yong~Jae Lee. 2024{\natexlab{a}}.
\newblock Improved baselines with visual instruction tuning.
\newblock In \emph{Proceedings of the IEEE/CVF Conference on Computer Vision and Pattern Recognition}, pages 26296--26306.

\bibitem[{Liu et~al.(2024{\natexlab{b}})Liu, Li, Li, Li, Zhang, Shen, and Lee}]{liu2024llavanext}
Haotian Liu, Chunyuan Li, Yuheng Li, Bo~Li, Yuanhan Zhang, Sheng Shen, and Yong~Jae Lee. 2024{\natexlab{b}}.
\newblock \href {https://llava-vl.github.io/blog/2024-01-30-llava-next/} {Llava-next: Improved reasoning, ocr, and world knowledge}.

\bibitem[{Liu et~al.(2024{\natexlab{c}})Liu, Li, Wu, and Lee}]{liu2024visual}
Haotian Liu, Chunyuan Li, Qingyang Wu, and Yong~Jae Lee. 2024{\natexlab{c}}.
\newblock Visual instruction tuning.
\newblock \emph{Advances in neural information processing systems}, 36.

\bibitem[{Liu et~al.(2025)Liu, Zheng, and Chen}]{liu2025paying}
Shi Liu, Kecheng Zheng, and Wei Chen. 2025.
\newblock Paying more attention to image: A training-free method for alleviating hallucination in lvlms.
\newblock In \emph{European Conference on Computer Vision}, pages 125--140. Springer.

\bibitem[{Maaz et~al.(2024)Maaz, Rasheed, Shaker, Khan, Cholakkal, Anwer, Baldwin, Felsberg, and Khan}]{maaz_palo_2024}
Muhammad Maaz, Hanoona~Abdul Rasheed, Abdelrahman~M. Shaker, Salman~H. Khan, Hisham Cholakkal, Rao~Muhammad Anwer, Tim Baldwin, Michael Felsberg, and Fahad~Shahbaz Khan. 2024.
\newblock \href {https://doi.org/10.48550/ARXIV.2402.14818} {{PALO}: {A} {Polyglot} {Large} {Multimodal} {Model} for {5B} {People}}.
\newblock \emph{CoRR}, abs/2402.14818.
\newblock ArXiv: 2402.14818.

\bibitem[{Nostalgebraist(2020)}]{nostalgebraist2020logitlens}
Nostalgebraist. 2020.
\newblock \href {https://www.lesswrong.com/posts/AcKRB8wDpdaN6v6ru/interpreting-gpt-the-logit-lens} {Interpreting gpt: The logit lens}.
\newblock LessWrong.

\bibitem[{OpenAI(2023)}]{gpt4}
OpenAI. 2023.
\newblock {GPT-4.}
\newblock \url{https://openai.com/gpt-4}.

\bibitem[{Pedregosa et~al.(2011)Pedregosa, Varoquaux, Gramfort, Michel, Thirion, Grisel, Blondel, Prettenhofer, Weiss, Dubourg, Vanderplas, Passos, Cournapeau, Brucher, Perrot, and Duchesnay}]{scikit-learn}
F.~Pedregosa, G.~Varoquaux, A.~Gramfort, V.~Michel, B.~Thirion, O.~Grisel, M.~Blondel, P.~Prettenhofer, R.~Weiss, V.~Dubourg, J.~Vanderplas, A.~Passos, D.~Cournapeau, M.~Brucher, M.~Perrot, and E.~Duchesnay. 2011.
\newblock Scikit-learn: Machine learning in {P}ython.
\newblock \emph{Journal of Machine Learning Research}, 12:2825--2830.

\bibitem[{Qin et~al.(2025)Qin, Chen, Zhou, Chen, Li, Liao, Li, Che, and Yu}]{qin2025survey}
Libo Qin, Qiguang Chen, Yuhang Zhou, Zhi Chen, Yinghui Li, Lizi Liao, Min Li, Wanxiang Che, and Philip~S Yu. 2025.
\newblock A survey of multilingual large language models.
\newblock \emph{Patterns}, 6(1).

\bibitem[{Qu et~al.(2024)Qu, Song, Wei, Dong, and Cheng}]{qu2024mitigating}
Xiaoye Qu, Mingyang Song, Wei Wei, Jianfeng Dong, and Yu~Cheng. 2024.
\newblock Mitigating multilingual hallucination in large vision-language models.
\newblock \emph{arXiv preprint arXiv:2408.00550}.

\bibitem[{Radford et~al.(2021)Radford, Kim, Hallacy, Ramesh, Goh, Agarwal, Sastry, Askell, Mishkin, Clark et~al.}]{radford2021learning}
Alec Radford, Jong~Wook Kim, Chris Hallacy, Aditya Ramesh, Gabriel Goh, Sandhini Agarwal, Girish Sastry, Amanda Askell, Pamela Mishkin, Jack Clark, et~al. 2021.
\newblock Learning transferable visual models from natural language supervision.
\newblock In \emph{International conference on machine learning}, pages 8748--8763. PMLR.

\bibitem[{Romero et~al.(2024)Romero, Lyu, Wibowo, Lynn, Hamed, Kishore, Mandal, Dragonetti, Abzaliev, Tonja, Balcha et~al.}]{romero_cvqa_2024}
David Romero, Chenyang Lyu, Haryo~Akbarianto Wibowo, Teresa Lynn, Injy Hamed, Aditya~Nanda Kishore, Aishik Mandal, Alina Dragonetti, Artem Abzaliev, Atnafu~Lambebo Tonja, Bontu~Fufa Balcha, et~al. 2024.
\newblock \href {https://doi.org/10.48550/ARXIV.2406.05967} {{CVQA}: {Culturally}-diverse {Multilingual} {Visual} {Question} {Answering} {Benchmark}}.
\newblock \emph{CoRR}, abs/2406.05967.
\newblock ArXiv: 2406.05967.

\bibitem[{S. et~al.(2023)S., C., S., K., X., T., and E.}]{Yin2023b}
Yin S., Fu~C., Zhao S., Li~K., Sun X., Xu~T., and Chen E. 2023.
\newblock A survey on multimodal large language models.
\newblock \emph{arXiv preprint arXiv:2306.13549}.

\bibitem[{Schneider and Sitaram(2024)}]{schneider_m5_2024}
Florian Schneider and Sunayana Sitaram. 2024.
\newblock \href {https://aclanthology.org/2024.findings-emnlp.250} {M5 - {A} {Diverse} {Benchmark} to {Assess} the {Performance} of {Large} {Multimodal} {Models} {Across} {Multilingual} and {Multicultural} {Vision}-{Language} {Tasks}}.
\newblock In \emph{Findings of the {Association} for {Computational} {Linguistics}: {EMNLP} 2024, {Miami}, {Florida}, {USA}, {November} 12-16, 2024}, pages 4309--4345. Association for Computational Linguistics.

\bibitem[{Schwenk et~al.(2022)Schwenk, Khandelwal, Clark, Marino, and Mottaghi}]{schwenk2022okvqa}
Dustin Schwenk, Apoorv Khandelwal, Christopher Clark, Kenneth Marino, and Roozbeh Mottaghi. 2022.
\newblock A-okvqa: A benchmark for visual question answering using world knowledge.
\newblock In \emph{European conference on computer vision}, pages 146--162. Springer.

\bibitem[{Shah et~al.(2023)Shah, Osi{\'n}ski, Levine et~al.}]{shah2023lm}
Dhruv Shah, B{\l}a{\.z}ej Osi{\'n}ski, Sergey Levine, et~al. 2023.
\newblock Lm-nav: Robotic navigation with large pre-trained models of language, vision, and action.
\newblock In \emph{Conference on robot learning}, pages 492--504. PMLR.

\bibitem[{Touvron et~al.(2023{\natexlab{a}})Touvron, Lavril, Izacard, Martinet, Lachaux, Lacroix, Rozière, Goyal, Hambro, Azhar et~al.}]{Llama2023}
Hugo Touvron, Thibaut Lavril, Gautier Izacard, Xavier Martinet, Marie-Anne Lachaux, Timothée Lacroix, Baptiste Rozière, Naman Goyal, Eric Hambro, Faisal Azhar, et~al. 2023{\natexlab{a}}.
\newblock \href {https://arxiv.org/abs/2302.13971} {Llama: Open and efficient foundation language models}.
\newblock \emph{arXiv preprint}.

\bibitem[{Touvron et~al.(2023{\natexlab{b}})Touvron, Martin, Stone, Albert, Almahairi, Babaei, Bashlykov, Batra, Bhargava, Bhosale et~al.}]{touvron2023llama}
Hugo Touvron, Louis Martin, Kevin Stone, Peter Albert, Amjad Almahairi, Yasmine Babaei, Nikolay Bashlykov, Soumya Batra, Prajjwal Bhargava, Shruti Bhosale, et~al. 2023{\natexlab{b}}.
\newblock Llama 2: Open foundation and fine-tuned chat models.
\newblock \emph{arXiv preprint arXiv:2307.09288}.

\bibitem[{Vaswani(2017)}]{vaswani2017attention}
A~Vaswani. 2017.
\newblock Attention is all you need.
\newblock \emph{Advances in Neural Information Processing Systems}.

\bibitem[{Wendler et~al.(2024)Wendler, Veselovsky, Monea, and West}]{wendler2024llamas}
Chris Wendler, Veniamin Veselovsky, Giovanni Monea, and Robert West. 2024.
\newblock Do llamas work in english? on the latent language of multilingual transformers.
\newblock \emph{arXiv preprint arXiv:2402.10588}.

\bibitem[{Wu et~al.(2023)Wu, Gan, Chen, Wan, and Philip}]{wu2023multimodal}
Jiayang Wu, Wensheng Gan, Zefeng Chen, Shicheng Wan, and S~Yu Philip. 2023.
\newblock Multimodal large language models: A survey.
\newblock In \emph{2023 IEEE International Conference on Big Data (BigData)}, pages 2247--2256. IEEE.

\bibitem[{Ye et~al.(2023)Ye, Xu, Ye, Yan, Liu, Qian, Zhang, Huang, and Zhou}]{ye2023mplug}
Qinghao Ye, Haiyang Xu, Jiabo Ye, Ming Yan, Haowei Liu, Qi~Qian, Ji~Zhang, Fei Huang, and Jingren Zhou. 2023.
\newblock mplug-owl2: Revolutionizing multi-modal large language model with modality collaboration.
\newblock \emph{arXiv preprint arXiv:2311.04257}.

\bibitem[{Yin et~al.(2023)Yin, Fu, Zhao, Li, Sun, Xu, and Chen}]{yin2023survey}
Shukang Yin, Chaoyou Fu, Sirui Zhao, Ke~Li, Xing Sun, Tong Xu, and Enhong Chen. 2023.
\newblock A survey on multimodal large language models.
\newblock \emph{arXiv preprint arXiv:2306.13549}.

\bibitem[{Yu et~al.(2024)Yu, Zhang, Yao, Dang, Chen, Lu, Cui, He, Liu, Chua et~al.}]{yu2024rlaif}
Tianyu Yu, Haoye Zhang, Yuan Yao, Yunkai Dang, Da~Chen, Xiaoman Lu, Ganqu Cui, Taiwen He, Zhiyuan Liu, Tat-Seng Chua, et~al. 2024.
\newblock Rlaif-v: Aligning mllms through open-source ai feedback for super gpt-4v trustworthiness.
\newblock \emph{arXiv preprint arXiv:2405.17220}.

\bibitem[{Z. et~al.(2024)Z., P., T., T., Z., Z., and MZ.}]{Bai2024}
Bai Z., Wang P., Xiao T., He~T., Han Z., Zhang Z., and Shou MZ. 2024.
\newblock Hallucination of multimodal large language models: A survey.
\newblock \emph{arXiv preprint arXiv:2404.18930}.

\bibitem[{Zhang et~al.(2024)Zhang, Huang, Jin, and Lu}]{zhang2024vision}
Jingyi Zhang, Jiaxing Huang, Sheng Jin, and Shijian Lu. 2024.
\newblock Vision-language models for vision tasks: A survey.
\newblock \emph{IEEE Transactions on Pattern Analysis and Machine Intelligence}.

\bibitem[{Zhao et~al.(2023)Zhao, Wang, Ouyang, Dong, Wang, and He}]{zhao2023hallucinations}
Zhiyuan Zhao, Bin Wang, Linke Ouyang, Xiaoyi Dong, Jiaqi Wang, and Conghui He. 2023.
\newblock \href {https://arxiv.org/abs/2311.16839} {Beyond hallucinations: Enhancing lvlms through hallucination-aware direct preference optimization}.
\newblock \emph{Preprint}, arXiv:2311.16839.

\bibitem[{Zhong et~al.(2024)Zhong, Feng, Zhao, Li, Huang, Gu, Ma, Xu, and Qin}]{zhong2024investigating}
Weihong Zhong, Xiaocheng Feng, Liang Zhao, Qiming Li, Lei Huang, Yuxuan Gu, Weitao Ma, Yuan Xu, and Bing Qin. 2024.
\newblock Investigating and mitigating the multimodal hallucination snowballing in large vision-language models.
\newblock \emph{arXiv preprint arXiv:2407.00569}.

\bibitem[{Zhu et~al.(2023)Zhu, Wang, Zhu, Sun, Zheng, Wang, and Chen}]{zhu2023prompt}
Peipei Zhu, Xiao Wang, Lin Zhu, Zhenglong Sun, Wei-Shi Zheng, Yaowei Wang, and Changwen Chen. 2023.
\newblock Prompt-based learning for unpaired image captioning.
\newblock \emph{IEEE Transactions on Multimedia}, 26:379--393.

\end{thebibliography}

\clearpage
\appendix
\section{Implementation Details}
\label{sec:implementation_details}

In our experiments, we use greedy search to ensure reproducibility.
We employ the experimental settings as default in the \href{https://github.com/DAMO-NLP-SG/VCD}{VCD} code repository.
All datasets used in this paper are licensed under a \href{https://creativecommons.org/licenses/by/4.0/legalcode}{Creative Commons Attribution 4.0 License}.
We conduct extensive experiments on languages exhibiting varying performance levels across the two models.
Probe is implemented as a linear Support Vector Machine (SVM) \cite{cortes1995support}, using the default \texttt{LinearSVC} API from \texttt{Scikit-learn} \cite{scikit-learn}.
When calculating metrics for discriminative tasks, we consider "yes" and "no" as right and wrong labels. The output of LVLMs are considered correct if its meaning matches the label.
We utilize ChatGPT \cite{gpt4} to assist us with coding and polishing the paper.

The hyperparameters under consideration included $\alpha$ and $K$.
The hyperparameter tuning strategy employed in this study follows a sequential optimization approach. 
We conduct hyperparameter tuning exclusively on the popular subset of POPE-COCO, with the search space for $K$ defined as \{50, 100, 150, 200, 250, 300\} and for $\alpha$ as \{0.5, 1.0, 1.5, 2.0, 2.5, 3.0, 3.5, 4.0, 4.5\}.
Initially, $K$ was fixed at a value of 100, while $\alpha$ was systematically adjusted to identify its optimal setting. Through this process, the optimal value for $\alpha$ was determined. 
Subsequently, $\alpha$ was fixed at this optimal value, and $K$ was iteratively tuned. This led to the identification of the optimal value for $K$. 
Consequently, the optimal hyperparameter combination was established as shown in Figure \ref{fig:ablation_alpha}. 
This stepwise optimization strategy ensures a focused and efficient exploration of the hyperparameter space, leading to the identification of the most effective parameter configuration for the model.

\section{Evaluation of Translation Quality}
\label{sec:trans}

We sample 100 translated queries from the POPE-COCO random subset for each language and back-translate them into English using Google Translate. The back-translated English queries are then input into LVLMs to test whether their predictions align with those generated from the original English queries. High prediction consistency indicates that the translated data maintains superior benchmark quality. 
Prediction consistency: Zh-100\%, Es-100\%, Ru-100\%, Pt-100\%, Bg-100\%, Hi-100\%, De-100\%.
These results demonstrate the reliability of our constructed multilingual dataset.

\section{Results of LLaVA-NeXT}
\label{sec:llava-next}
To comprehensively demonstrate the effectiveness of \ours, we also conduct experiments on an advanced LVLM, LLaVA-NeXT-7b \cite{liu2024llavanext}, with stronger English capability than LLaVA-1.5-7b, as shown in Table \ref{tab:POPE_next}, also demonstrating strong effectiveness.

\begin{table*}[!ht]
\centering
\setlength\tabcolsep{2pt} 
\begin{tabular}{lllccccccc}
\toprule
\textbf{Dataset}& \textbf{Setup} & \textbf{Method}& \textbf{En} & \textbf{Zh} & \textbf{Es} & \textbf{Ru} & \textbf{Pt} & \textbf{Bg} &\textbf{Avg.}\\
\midrule
\multirow{9}{*}{\makecell{COCO}} & \multirow{3}{*}{Random} & \textit{Baseline} & 88.50&88.33&80.30&88.37&86.77&81.70&85.09 \\
 &  & \textit{VCD} & -& 85.83&69.47&84.53&81.97&75.67&79.49 \\
&  & \textit{Ours} & -& 88.73&88.47&88.13&86.73&81.67&86.75  \\
  \cmidrule(lr){3-10} 
  & \multirow{3}{*}{Popular} & \textit{Baseline} & 87.37&87.53&80.07&85.60&85.13&79.30&83.53	\\
 &  & \textit{VCD} & -&85.43&69.20&82.07&82.93&75.67&79.06 \\
&  & \textit{Ours} & -&88.43&88.13&85.83&85.40&80.33&85.62  \\
  \cmidrule(lr){3-10} 
  & \multirow{3}{*}{Adversarial} & \textit{Baseline} & 86.30&80.80&79.20&80.60&80.37&76.10&79.41 \\
 &  & \textit{VCD} & -& 77.23&68.10&76.93&76.27&70.30&73.77 \\
&  & \textit{Ours} & -& 80.63&83.83&80.20&80.30&75.83&80.16  \\
 \midrule
 
 \multirow{9}{*}{\makecell{OKVQA}} & \multirow{3}{*}{Random} & \textit{Baseline} & 91.00&83.03&67.50&85.00&78.97&81.20&79.14 \\
 &  & \textit{VCD} & -&82.20&69.27&82.23&79.37&75.70&77.75\\
&  & \textit{Ours} & -&85.47&87.50&86.50&83.33&80.90&84.74  \\
  \cmidrule(lr){3-10} 
  & \multirow{3}{*}{Popular} & \textit{Baseline} & 89.00&81.93&67.40&80.73&80.80&79.90&78.15	 \\
 &  & \textit{VCD} & -&81.23&69.33&78.33&79.77&74.27&76.59 \\
&  & \textit{Ours} & -&85.33&85.27&82.13&85.07&80.13&83.59  \\
  \cmidrule(lr){3-10} 
  & \multirow{3}{*}{Adversarial} & \textit{Baseline} & 81.97&70.80&66.47&74.17&71.27&73.33&71.21	 \\
 &  & \textit{VCD} & -&71.57&66.97&71.67&70.83&69.63&70.13	 \\
&  & \textit{Ours} & -&75.40&76.93&76.73&75.17&73.70&75.59  \\
 \midrule
 \multirow{9}{*}{\makecell{GQA}} & \multirow{3}{*}{Random} & \textit{Baseline} & 89.93&82.53&67.73&84.47&80.07&79.87&78.93\\
 &  & \textit{VCD} & -&81.03&68.60&81.57&77.63&74.53&76.67\\
&  & \textit{Ours} & -&84.23&87.13&85.47&83.30&79.97&84.02\\
  \cmidrule(lr){3-10} 
  & \multirow{3}{*}{Popular} & \textit{Baseline} & 85.97&77.17&67.70&73.93&74.90&79.73&74.69 	 \\
 &  & \textit{VCD} & -&76.80&69.53&73.60&75.23&74.20&73.87	\\
&  & \textit{Ours} & -&80.50&83.77&76.50&81.83&81.03&80.73  \\
  \cmidrule(lr){3-10} 
  & \multirow{3}{*}{Adversarial} & \textit{Baseline} & 82.60&70.33&66.93&73.90&71.50&74.03&71.34 \\
 &  & \textit{VCD} & -&69.57&67.50&72.90&71.57&69.33&70.17 \\
&  & \textit{Ours} & -&74.07&78.10&76.93&76.53&74.87&76.10  \\
\bottomrule
\end{tabular}
\caption{Main results of LLaVA-NeXT on POPE from COCO, OKVQA, GQA. 
}
\label{tab:POPE_next}
\end{table*}

\section{Inference Speed}
\label{sec:latency}
We evaluate Tokens Per Second (TPS) of LLaVA-1.5 on the popular subset of POPE-COCO using different methods, with the experiments conducted on the H100 GPUs. 
The results show that VCD leads to a significant decrease in inference speed. 
We attribute this slowdown to the fact that contrastive-decoding-based methods typically require multiple inference runs or involve substantial additional computations during the inference process. 
In contrast, \ours introduces almost no extra computational overhead during inference, further showing the advantages of our method.

\vspace{2mm}
\begin{tabular}{llc}
\toprule
Method & TPS & Acc (\%)\\
\midrule
LLaVA-1.5-7b & 55.46 \textcolor{gray}{\(\times\)1.0} & 73.68 \\
+VCD & 20.67 \textcolor{red}{\(\times\)0.4} & 74.03 \\
\midrule
+Ours & 54.79 \textcolor[rgb]{0,0.5,0}{\(\times\)1.0}  & 85.15 \\
\bottomrule
\end{tabular}

\section{Detailed Results on MME}
\label{sec:mme_full}
\begin{table*}[!ht]
\centering
\small 
\setlength\tabcolsep{2pt} 
\fontsize{9}{10}\selectfont
\begin{tabularx}{\textwidth}{llCCCCCCCCCCCCCC}
\toprule
\multirow{2}{*}{\textbf{Task}} & \multirow{2}{*}{\textbf{Method}}& \multicolumn{7}{c}{\textbf{LLaVA-1.5}} & \multicolumn{7}{c}{\textbf{Qwen-VL-Chat}} \\
\cmidrule(lr){3-9} \cmidrule(lr){10-16}
&  & \textbf{En} & \textbf{Zh} & \textbf{Es} & \textbf{Ru} & \textbf{Pt} & \textbf{Bg} &\textbf{Avg.}& \textbf{En}&\textbf{Zh} & \textbf{Es} & \textbf{Ru} & \textbf{Hi} & \textbf{De}&\textbf{Avg.}\\
\midrule
\multirow{3}{*}{Existence}&\textit{Baseline}&190.0&175.0&130.0&145.0&155.0&125.0&146.0&185.0&190.0&135.0&140.0&106.7&195.0&153.3\\
&\textit{VCD}&-&180.0&130.0&145.0&150.0&128.3&146.7&-&185.0&155.0&155.0&78.30&195.0&153.7\\
&\textit{Ours}&-&195.0&175.0&185.0&175.0&180.0&182.0&-&190.0&155.0&185.0&128.3&195.0&170.7\\
\cmidrule(lr){2-16} 
\multirow{3}{*}{Count}&\textit{Baseline}&155.0&70.00&55.00&58.30&80.00&85.00&69.67&150.0&130.0&143.3&113.3&60.00&136.7&116.7\\
&\textit{VCD}&-&73.30&73.30&53.30&61.70&105.0&73.33&-&140.0&131.7&100.0&80.00&128.3&116.0\\
&\textit{Ours}&-&125.0&130.0&110.0&130.0&135.0&126.0&-&148.3&153.3&120.0&111.7&137.0&134.0\\
\cmidrule(lr){2-16} 
\multirow{3}{*}{Color}&\textit{Baseline}&165.0&80.00&135.0&75.00&110.0&80.00&96.00&180.0&170.0&150.0&153.3&103.3&165.0&148.3\\
&\textit{VCD}&-&95.00&145.0&85.00&130.0&88.30&108.7&-&150.0&165.0&146.7&93.30&160.0&143.7\\
&\textit{Ours}&-&120.0&155.0&125.0&150.0&108.3&131.7&-&170.0&165.0&158.3&90.00&165.0&149.7\\
\cmidrule(lr){2-16} 
\multirow{3}{*}{Position}&\textit{Baseline}&118.3&53.30&63.30&55.00&50.00&46.70&53.67&131.7&63.30&120.0&93.30&45.00&105.0&85.33\\
&\textit{VCD}&-&48.30&93.00&60.00&51.70&56.70&61.93&-&78.30&101.7&101.7&43.30&93.30&83.67\\
&\textit{Ours}&-&66.70&78.30&85.00&86.70&58.30&75.00&-&63.30&116.7&103.3&55.00&106.0&88.86\\
\midrule
\multirow{3}{*}{Artwork} & \textit{Baseline} & 121.8&86.25&57.75&73.00&81.25&61.25&71.90&135.3&152.8&105.8&70.00&65.75&130.5&105.0 \\
& \textit{VCD} & -& 96.75&79.50&81.75&74.50&72.25&80.95&-&137.5&108.5&71.75&55.50&123.3&99.30\\
& \textit{Ours} & -& 117.5&118.0&103.5&107.3&98.25&108.9&-&146.0& 110.0&115.0&83.00&133.8&117.6 \\
\cmidrule(lr){2-16}
\multirow{3}{*}{Celebrity} & \textit{Baseline} & 138.2&137.4&55.29&26.18&119.7&13.24&70.35&150.0&162.9&159.4&52.65&57.65&126.8&111.9	\\
& \textit{VCD} & -&139.7&79.12&23.82&99.41&10.59&70.53&-&160.9&158.5&67.35&52.35&121.2&112.1 \\
& \textit{Ours} & -&156.5&158.2&61.47&145.3&27.65&109.8&-&157.9&160.3&110.6&76.47&106.2&122.3  \\
\cmidrule(lr){2-16} 
\multirow{3}{*}{OCR} & \textit{Baseline} & 125.0&65.00&50.00&55.00&15.00&55.00&48.00&102.5&80.00&115.0&62.50&32.50&132.5&84.50 \\
& \textit{VCD} & -& 62.50&57.50&52.50&35.00&57.50&53.00&-&87.50&100.0&52.50&42.50&115.0&79.50 \\
& \textit{Ours} & -& 80.00&82.50&57.50&32.50&65.00&63.50&-&87.50&100.0&77.50&50.00&135.0&90.00  \\
 \cmidrule(lr){2-16}
 
\multirow{3}{*}{Landmark} & \textit{Baseline} & 165.3&113.3&53.75&81.50&127.3&108.5&96.90&172.8&179.0&123.3&78.25&61.50&163.3&121.1	 \\
& \textit{VCD} & -&125.5&64.25&97.25&126.3&108.0&102.3&-&167.0&131.3&78.25&57.00&138.0&114.3\\
& \textit{Ours} & -&147.8&155.5&146.5& 159.3&148.3&151.5&-&178.3&129.8&160.8&88.25&167.5&144.9 \\
  \cmidrule(lr){2-16} 
\multirow{3}{*}{Scene} & \textit{Baseline} & 159.5&159.3&79.50&120.0&144.3&122.5&125.1&161.5&178.8&124.8&114.5&72.75&131.5&124.5	 \\
& \textit{VCD} & -&149.8&87.30&119.5&137.5&118.0&122.4&-&162.3&122.8&108.0&71.50&128.8&118.7 \\
& \textit{Ours} & -&149.5&146.8&154.3&144.3&145.0&148.0&-&162.5&123.5&153.0&92.50&143.5&135.0  \\
\cmidrule(lr){2-16}
\multirow{3}{*}{Poster} & \textit{Baseline} & 143.5&106.1&67.35&57.14&110.2&68.71&81.90&173.1&156.8&142.9&116.7&76.19&134.4&125.4	 \\
& \textit{VCD} & -&116.7&89.80&65.31&90.48&80.95&88.64&-&147.3&137.4&101.0&65.99&126.9&115.7 \\
& \textit{Ours} & -&133.7&156.8&97.62&129.9&121.8&128.0&-&161.6&150.3&138.8&109.9&137.8&139.7  \\
\midrule
\multirow{3}{*}{Code Reasoning} & \textit{Baseline} & 67.50&65.00&50.00&47.50&55.00&22.50&48.00&55.00&57.50&57.50&20.00&12.50&52.50&40.00 	 \\
& \textit{VCD} & -&67.50&57.50&72.50&62.50&27.50&57.50&-&42.50&57.50&37.50&35.00&45.00&43.50 	\\
& \textit{Ours} & -&55.00&60.00&47.50&65.00&25.00&50.50&-&72.50&50.00&45.00&10.00&52.50&46.00  \\
  \cmidrule(lr){2-16} 
\multirow{3}{*}{Numerical Calculation} & \textit{Baseline} & 70.00&47.50&50.00&45.00&20.00&20.00&36.50&32.50&65.00&45.00&27.50&45.00&37.50&44.00 \\
& \textit{VCD} & -&75.00&50.00&62.50&37.50&20.00&49.00&-&80.00&60.00&37.50&45.00&45.00&53.50 \\
& \textit{Ours} & -&67.50&50.00&55.00&20.00&30.00&44.50&-&45.00&45.00&55.00&45.00&27.50&43.50  \\
  \cmidrule(lr){2-16} 
\multirow{3}{*}{Text Translation} & \textit{Baseline} & 70.00&77.50&50.00&70.00&120.0&52.50&74.00&155.0&60.00&115.0&110.0&17.50&65.00&73.50 \\
& \textit{VCD} & -&80.00&50.00&72.50&75.00&72.50&70.00&-&55.00&87.50&85.00&45.00&72.50&69.00 \\
& \textit{Ours} & -&95.00&57.50&50.00&90.00&57.50&70.00&-&87.50&122.5&110.0&50.00&72.50&88.50  \\
  \cmidrule(lr){2-16} 
\multirow{3}{*}{Commonsense Reasoning} & \textit{Baseline} & 124.3&72.14&64.29&61.43&72.86&64.29&67.00&125.0&92.14&100.7&50.71&33.57&74.29&70.28 \\
& \textit{VCD} & -&80.71&71.43&70.00&77.14&62.86&72.43&-&84.29&98.57&48.57&29.29&78.57&67.86 \\
& \textit{Ours} & -&90.00&87.14&86.43&87.14&68.57&85.86&-&91.43&107.1&80.00&39.29&84.29&80.43  \\
\midrule
\multirow{3}{*}{Total Scores} & \textit{Baseline} & 1813&1307&961.3&970.1&1261&925.2&1085&1909&1738&1638&1203&790.0&1650&1404 \\
& \textit{VCD} & -&1391&1128&1051&1209&1009&1157&-&1683&1619&1191&794.1&1571&1371\\
& \textit{Ours} & -&1599&1611&1375&1522&1269&1475&-&1762&1689&1612&1029&1664&1551\\

\bottomrule
\end{tabularx}
\caption{Detailed results on full subsets of MME. 
}
\label{tab:mme_all}
\end{table*}
In Table~\ref{tab:mme_all}, we present the performance of LVLM baselines on the 14 tasks of the MME benchmark. 
The Existence and Count subsets assess object-level hallucination, while the Color and Position subsets focus on attribute-level hallucination. These four subsets form the hallucination evaluation set of MME, which, together with the remaining six categories—Artwork, Celebrity, OCR, Landmark, Scene, and Poster—collectively evaluate the LVLMs' perception capabilities.
Besides, Code Reasoning, Numerical Calculation, Text Translation and Commonsense Reasoning evaluate the LVLMs' recognition capabilities.
The deployment of \ours nearly consistently improves their perceptual competencies, indicating the effectiveness of \ours for cross-lingual visual perception alignment. 
Furthermore, The results on recognition-related tasks indicate that the application of \ours, while mitigating hallucination issues and augmenting perceptual capabilities, remains effective on some reasoning tests.
\section{Impact of Training Size}
\label{sec:size}
\begin{figure}[!h]
  \includegraphics[width=\columnwidth]{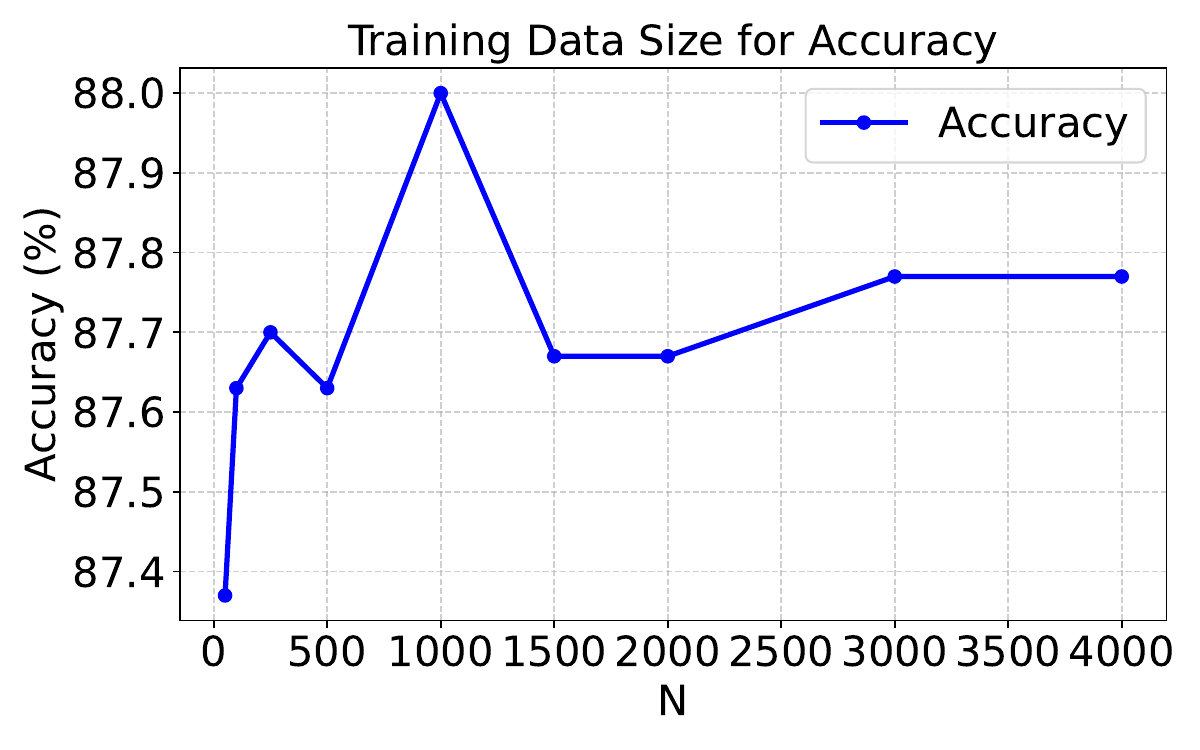}
  \caption{Impact of training data size on the Accuracy for LLaVA-1.5.}
  \label{fig:n}
\end{figure}

In the main experiment, we sample 1,000 images from the COCO-2017 training dataset to identify the \headname and estimate \shift.
Of these, 80\% are used as the training set for the probe, while the remaining 20\% serve as the test set for the probe.
Specifically, \ours achieves strong performance even with a minimal training set. 
We validate this on the popular subset of POPE-COCO (Chinese), as shown in Figure \ref{fig:n}. 
Notably, most improvements are achieved with as few as N = 50 training samples. 
As the training set size increases, identification becomes more robust; however, an excessively large training set may introduce additional noise when estimating the mean of the \shift.
\section{Comparison to PAI}
According to the suggestion of the anonymous reviewer, we conduct a comparison experiment with PAI \cite{liu2025paying}, which intervenes on attention heads by leveraging their original direction. In Table \ref{tab:pai}, CLAIM still achieves best performance.
\section{Case Study}
\begin{table*}[!h]
\centering
\setlength\tabcolsep{4pt} 
\begin{tabular}{llcccccc}
\toprule
\textbf{Setup} & \textbf{Method}& \textbf{Zh} & \textbf{Es} & \textbf{Ru} & \textbf{Pt} & \textbf{Bg} &\textbf{Avg.}\\
\midrule
\multirow{4}{*}{Random} & \textit{Baseline} & 81.00&63.03&72.33&78.97&72.23&73.51 \\
  & \textit{VCD} &  81.47&67.40&73.33&78.07&72.47&74.55 \\
  & \textit{PAI} & 80.10&67.97&83.50&79.13&71.10&76.36\\
  & \textit{Ours} & 86.50&87.33&87.50&85.40&80.50&85.45  \\
  \cmidrule(lr){3-8} 
   \multirow{4}{*}{Popular} & \textit{Baseline} & 83.07&62.93&69.20&82.03&71.17&73.68	\\
   & \textit{VCD} & 83.20&67.20&70.17&80.10&69.47&74.03 \\
   & \textit{PAI} & 83.90&67.73&79.17&82.90&71.90&77.12\\
  & \textit{Ours} & 88.00&87.53&84.03&86.13&80.07&85.15  \\
  \cmidrule(lr){3-8} 
   \multirow{4}{*}{Adversarial} & \textit{Baseline} & 73.40&62.87&66.07&73.70&65.93&68.39 \\
   & \textit{VCD} &  74.27&67.30&66.47&73.63&66.33&69.60 \\
   & \textit{PAI} & 76.60&67.43&75.20&75.53&67.73&72.50\\
  & \textit{Ours} &  77.67&83.27&79.43&79.67&74.57&78.92  \\
\bottomrule
\end{tabular}
\caption{Comparison to PAI on POPE of LLaVA-1.5. 
}
\label{tab:pai}
\end{table*}
\begin{figure*}[!h]
  \includegraphics[width=\textwidth]{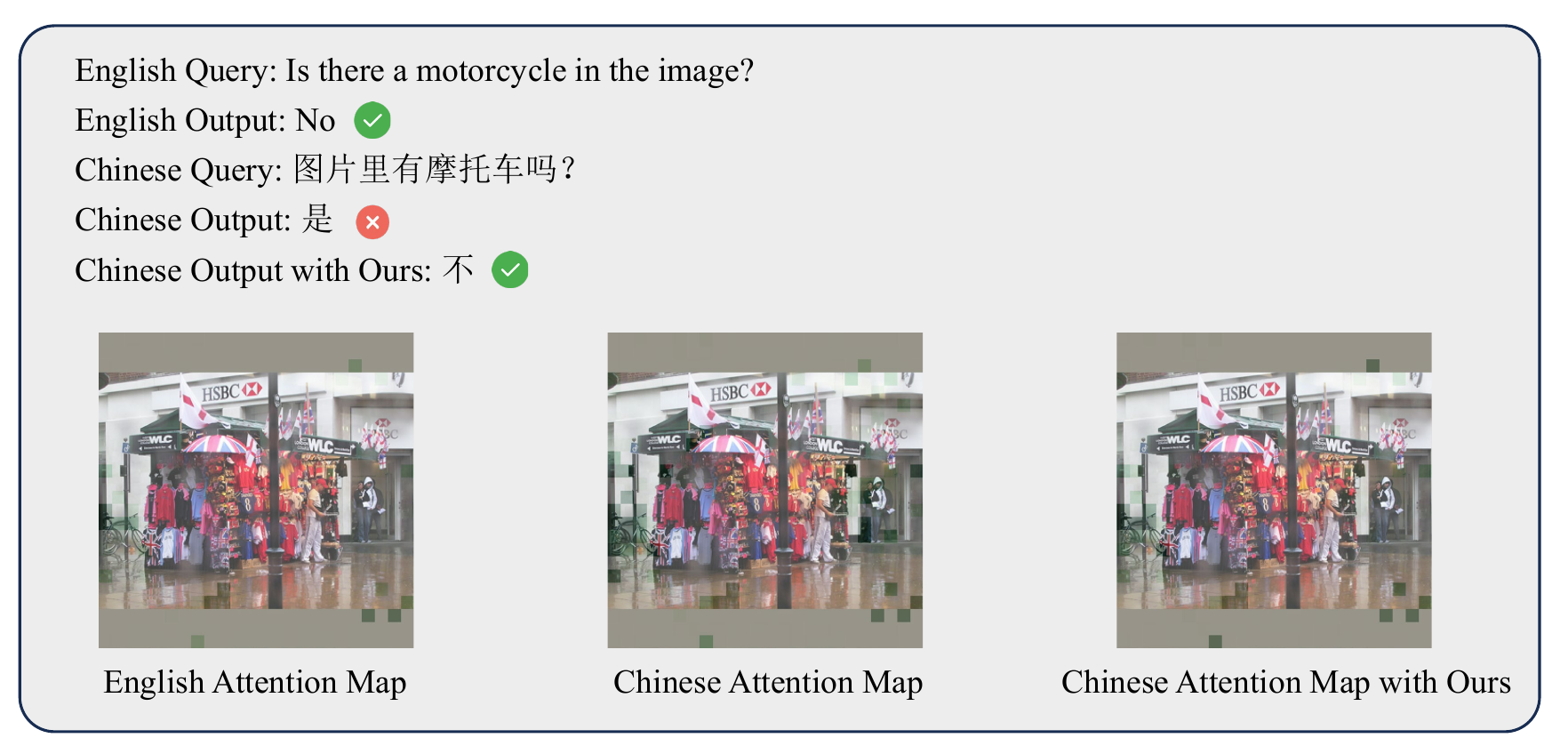}
  \includegraphics[width=\textwidth]{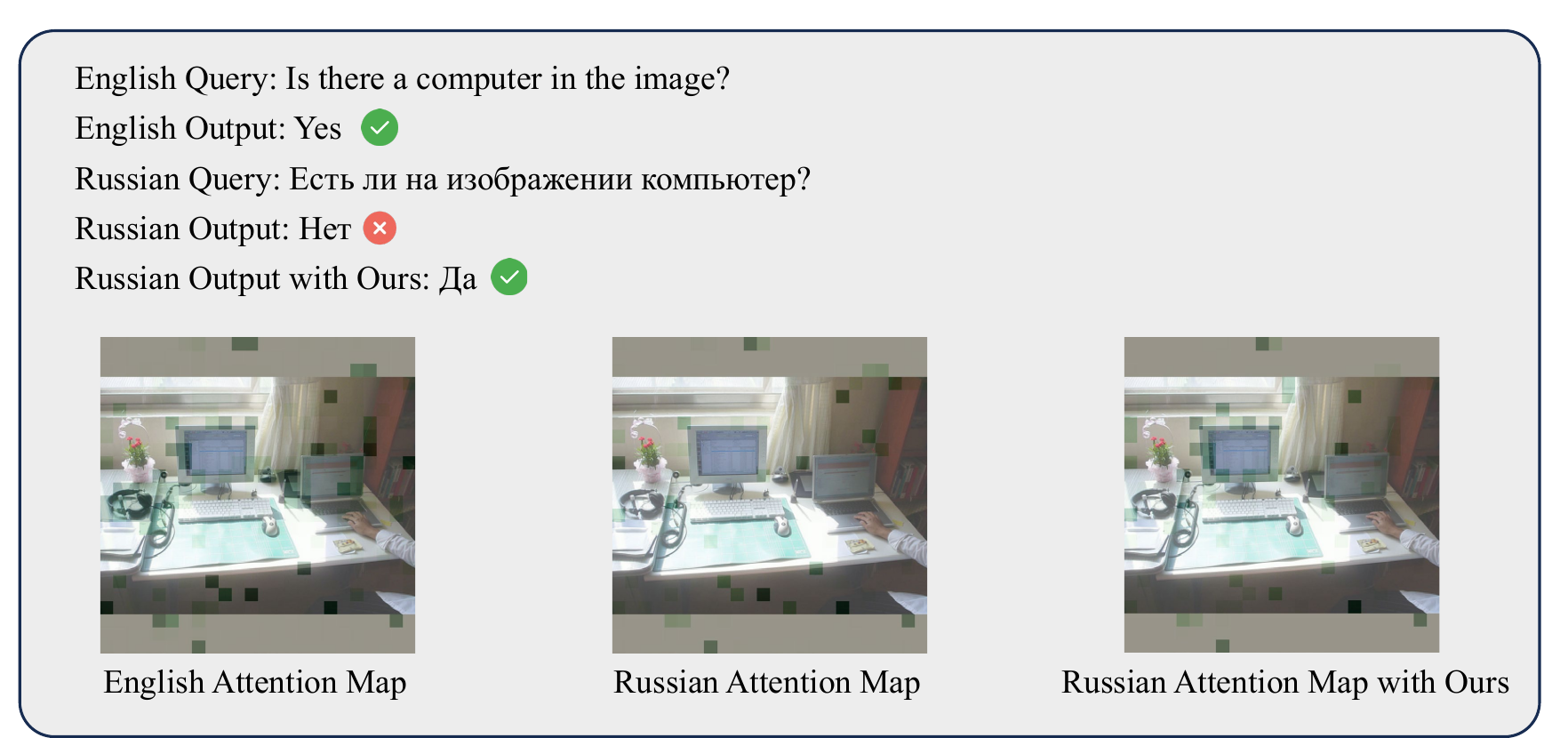}
  \caption{Case study.}
  \label{fig:case}
\end{figure*}
Illustrated in Figure \ref{fig:case}, we present examples demonstrating how CLAIM effectively mitigates multilingual object hallucination. 
We compare the changes in the model's attention weight maps for the same image before and after applying CLAIM.

\end{document}